\documentclass{article}

\PassOptionsToPackage{numbers,compress}{natbib}
\usepackage[main, final]{neurips_2026}

\usepackage[utf8]{inputenc}
\usepackage[T1]{fontenc}
\usepackage{hyperref}
\usepackage{url}
\usepackage{booktabs}
\usepackage{amsfonts}
\usepackage{nicefrac}
\usepackage{microtype}
\usepackage{xcolor}
\usepackage[ruled,linesnumbered]{algorithm2e}
\SetKwInput{Input}{Input}
\SetKwInput{Output}{Output}

\usepackage{graphicx}
\usepackage{wrapfig}
\usepackage{amsmath}
\usepackage{multirow}
\usepackage{array}
\usepackage{booktabs}
\usepackage{subcaption}
\usepackage{enumitem}
\newcommand{\sysname}{gHAWK}

\title{gHAWK: Structural Encoding for Scalable Training of Graph Neural Networks on Knowledge Graphs}

\author{Humera Sabir\thanks{Equal contribution.}
  \qquad
  Fatima Farooq\footnotemark[1]
  \qquad
  Ashraf Aboulnaga \\
  Department of Computer Science and Engineering\\
  University of Texas at Arlington\\
  Arlington, TX, USA\\
  \texttt{\{humera.sabir,fatima.farooq,ashraf.aboulnaga\}@uta.edu}
  }
\begin{document}
\maketitle
\vspace{-18pt}
\begin{abstract}
\vspace{-4pt}
Knowledge Graphs (KGs) are a rich source of structured data, and graph neural networks (GNNs) are the dominant tool for learning over them. However, existing message-passing GNNs struggle to scale to large KGs because they rely on the iterative message passing process to learn graph structure. This process is inefficient, especially under mini-batch training, where a node sees only a partial view of its neighborhood. In this paper, we address this problem and present \sysname{}, a novel and scalable GNN training framework for large KGs. The key idea is to precompute structural features for each node that capture its local and global structure before GNN training even begins. Specifically, \sysname{} introduces a preprocessing step that computes: (a)~Bloom filters to compactly encode local neighborhood structure, and (b)~TransE embeddings to represent each node's global position in the graph. These features are then fused with any domain-specific features (e.g., text embeddings), producing a node feature vector that can be used with any GNN backbone. Equipped with these structural priors, the GNN no longer needs to rediscover graph structure, significantly improving convergence and model accuracy with little memory overhead. Extensive experiments on large datasets from the Open Graph Benchmark demonstrate that \sysname{} improves accuracy on both node property prediction and link prediction tasks. Remarkably, \sysname{} enables simple GNNs to match or exceed more complex relation-aware ones, suggesting that careful preprocessing can substitute for architectural complexity in KG learning.
\end{abstract}

\vspace{-9pt}
\section{Introduction}
\vspace{-9pt}

Knowledge graphs (KGs) represent real-world entities and the relations among them as (head, relation, tail) triples. 
They support applications such as question answering~\cite{appQA1, appQA2, multihop}, recommendation systems~\cite{appRE1},
information retrieval~\cite{appIR1,appIR2}, and biomedical reasoning~\cite{KGBiomed}.
To leverage the rich, heterogeneous structure in a KG, a first step in many applications is to train a machine learning model that learns representations. 
Graph neural networks (GNNs) have emerged as a powerful class of models for this step, which learn vector representations of entities, also known as embeddings. 
At each layer, information from each entity's neighborhood is aggregated through \emph{message passing}~\cite{gilmer2017neural, rgcncloser}: every node updates its embedding by combining messages from its neighbors~\cite{RGCN, CompGCN, clustergcn}. By aggregating this structural information across multiple layers, GNNs improve predictive accuracy on downstream tasks such as node property prediction and link prediction.

While GNNs have demonstrated remarkable predictive accuracy, training them on large-scale KGs poses serious scalability challenges, and three properties of KGs make these challenges particularly acute.
First, KGs are heterogeneous with high structural diversity: a single graph may contain hundreds of distinct relation types~\cite{wikidata}, each carrying different semantics, which relation-aware GNNs encode through per-relation parameters~\cite{RGCN, CompGCN}. 
The parameter count grows linearly in the number of relations, increasing memory cost and slowing convergence. 
Second, full-batch training is infeasible at the scale of millions of entities, so training relies on mini-batches that sample only a fraction of each node's neighborhood per step~\cite{graphSAGE, fastgcn, graphsaint}. 
Sampling discards exactly the structural information the GNN is trying to learn. 
Third, message passing is local: at each layer, every node can only access information from its immediate neighbors, so capturing structure beyond one hop requires multiple message-passing iterations, and each iteration adds memory and compute cost.
Taken together, these three factors mean that \textbf{GNNs must reconstruct KG structure during training, one hop at a time, even though that structure is already explicit in the input graph.} 

We argue that 
GNN training on large
KGs requires \emph{precomputed structural priors}: features
computed once over the 
input graph that encode each entity's
structural context before 
GNN training starts.
Our insight is to \textbf{shift the complexity of
capturing KG structure from the message-passing stage to a
lightweight preprocessing step.} The design question becomes:
\emph{what structural information should this prior carry?} We show that two
complementary priors are sufficient. The \emph{local} feature
vector captures an entity's $1$-hop neighborhood, recovering
information that mini-batch sampling discards. The \emph{global}
feature vector captures the entity's position in graph,
providing relational context that message passing would
otherwise need many layers to propagate. Together, these
priors give each entity a representation of both its local and
distant structure, available before the first training step.

\begin{figure*}[t!]
  \centering
\includegraphics[width=0.95\textwidth]{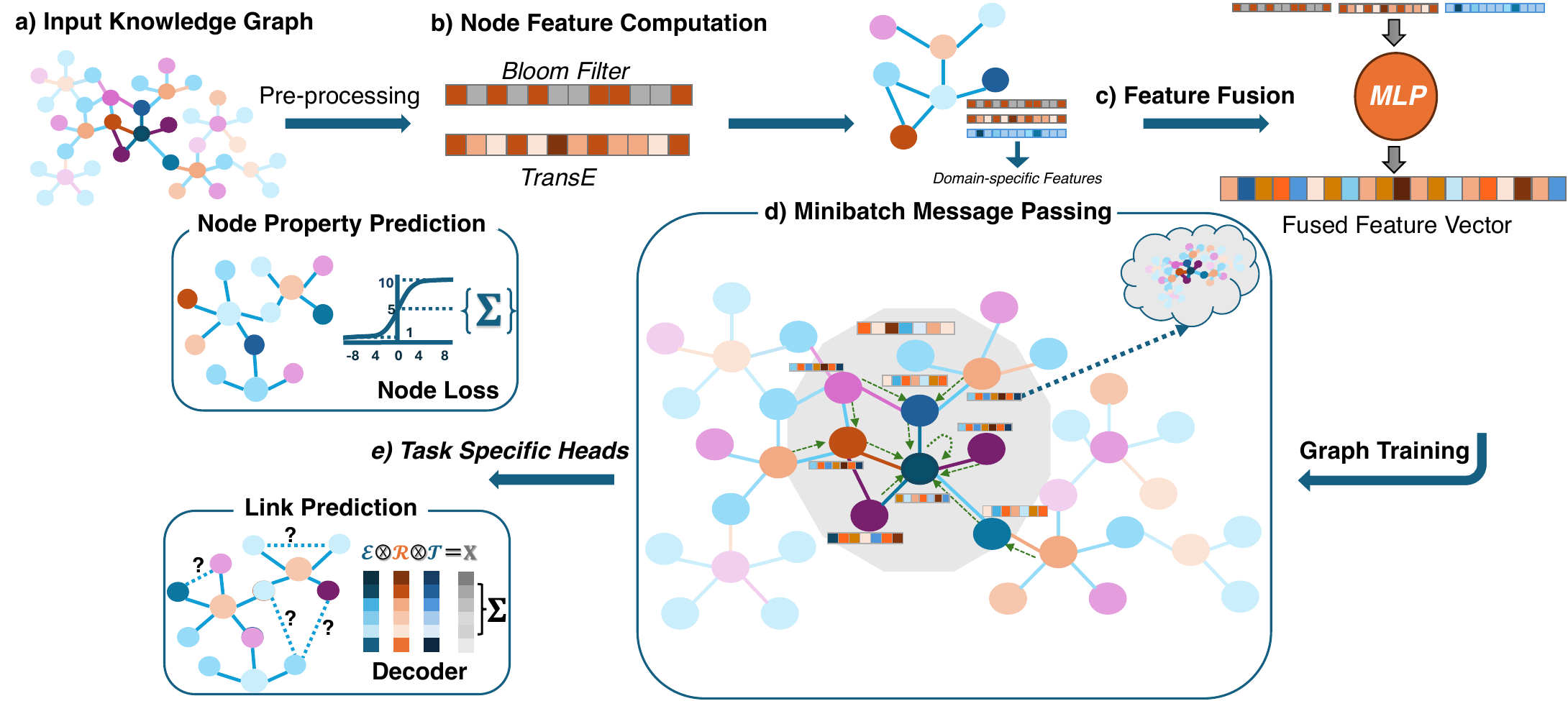}
  \caption{\sysname{} pipeline. (a)~Input knowledge graph.
  (b)~Node feature computation: in a preprocessing step, \sysname{} constructs a Bloom filter encoding the $1$-hop neighbors of each node and trains a TransE embedding model to capture the global graph structure. 
  (c)~Feature fusion: for every node, the Bloom filter, TransE embedding, and any domain-specific feature vector are fused via an MLP trained with task supervision, creating a dense fused feature vector. 
  (d)~Mini-batch message passing: the fused vectors initialize a message-passing GNN that is trained on sampled mini-batch subgraphs of the knowledge graph.
  (e)~Task-specific heads: the resulting node embeddings are fed into separate heads for node property prediction 
  and for link prediction.
  }
  \label{fig:model-arch}
\vspace{-9pt}
\end{figure*}

We instantiate this design in
\sysname{}\footnote{\url{https://github.com/lids-lab/gHAWK}}
(\emph{Hybrid
Aggregation of local and global structure for Web-scale KGs}),
a framework for KG learning that combines local and global
structural priors with any GNN architecture and any downstream
prediction task
(Fig.~\ref{fig:model-arch}). For the local feature,
\sysname{} uses a \emph{Bloom filter}~\cite{bloom-org} that
compactly encodes each entity's $1$-hop neighbors. For the global feature, \sysname{} uses a pre-trained knowledge graph
embedding (KGE) model, specifically \emph{TransE}~\cite{TransE}, that
captures each entity's global relational position. \sysname{} fuses these two structural
vectors with a learned entity embedding, and with any
domain-specific node features (e.g., text embeddings in
bibliographic KGs), through a learned gating mechanism. The
resulting dense vector serves as input either to a GNN backbone or directly to a task-specific
decoder.
The framework can accommodate alternative local priors (relation-distribution histograms, MinHash) or global priors (DistMult, ComplEx) and we show that they yield comparable performance. We adopt Bloom + TransE as default because both are simple to compute and provide a compact local signature and an interpretable positional encoding. While precomputed node features are not new, we are not aware of any prior work that combines local and global structural priors in this way to scale GNN training on large heterogeneous KGs. 

We show that \sysname{} improves accuracy on 
multiple benchmarks for two tasks: node property prediction and link prediction.
In summary, our contributions are:

\textbf{A scalable framework that decomposes structure into local and global priors.} \sysname{} computes two complementary fixed-size structural features for each entity, instantiated by default with a Bloom filter and a pre-trained TransE embedding. A learned gating mechanism fuses these frozen priors with the entity embedding, allowing the model to balance the precomputed structural signal against task-specific learning during training.

\textbf{Structural priors substitute for deep message passing across tasks and backbones.} Across both node property prediction and link prediction, simple architectures equipped with \sysname{} features match or surpass message-passing alternatives. On node property prediction over OGB-MAG and MAG240M~\cite{OGB, OGBLSC}, \sysname{} consistently improves accuracy across six GNN backbones spanning relation-aware, relation-agnostic, sampling-based, and attention-based architectures. On link prediction over OGB-WikiKG2, \sysname{} features with a RotatE decoder--using no message passing at all--improve test MRR by $26$ percentage points over the strongest embedding-only baseline and exceed specialized query-conditioned methods such as A\textsuperscript{$\star$}Net~\cite{astartnet}. On OGB-BioKG, a biomedical KG with type-constrained evaluation, \sysname{} improves over RotatE by up to 9.6 MRR points on clinically important relation types (drug--disease, drug--drug, drug--side-effect), where the base model is weakest. Stratified analysis traces both gains to \sysname{}'s effectiveness at capturing the long-tail: rare entities on WikiKG2 and difficult relations on BioKG. Where learned embeddings provide weak signal, structural priors restore predictive capacity.

\textbf{A theoretical understanding of the structural-prior paradigm.} Beyond gHAWK's empirical results, our analysis reveals a fundamental property of query-independent KG representations: certain entity pairs are structurally indistinguishable from any feature defined over neighborhoods alone, regardless of how the feature is constructed. We call these doppelgangers, and they create a ceiling for the accuracy of query-independent methods, which we analyze in Sec.~\ref{sec:exp-robustness} and App.~\ref{app:proofs}.

\vspace{-12pt}

\section{Background}
\vspace{-9pt}
\label{sec: background}

A knowledge graph (KG) is $G = (\mathcal{E}, \mathcal{R}, \mathcal{T})$ 
with entities $\mathcal{E}$, relation types $\mathcal{R}$, and 
triples $\mathcal{T} \subseteq \mathcal{E} \times \mathcal{R} \times \mathcal{E}$. 
Each triple $(h, r, t)$ is a directed edge from a head entity 
$h$ to a tail entity $t$ labeled with a relation type $r$. For 
an entity $e$, we write $\mathcal{N}(e) = \{(r, t) : (e, r, t) \in \mathcal{T}\} \cup 
\{(r, h) : (h, r, e) \in \mathcal{T}\}$ for its 1-hop 
neighborhood combining outgoing and incoming edges.

\vspace{-9pt}
\paragraph{Set signatures.}
\sysname{} encodes each entity's 1-hop neighborhood with a 
compact, fixed-size set signature, which lets every entity 
carry its full neighborhood through training as a constant-size 
vector regardless of how many neighbors it has. We consider two 
well-established structures. 
Our default is a \emph{Bloom filter}~\cite{bloom-org}, an $m$-bit array with $k$ independent hash functions; 
inserting an element sets $k$ bits, yielding a fixed-size 
signature with no false negatives and a tunable false-positive 
rate. We also test \emph{MinHash}~\cite{minhash}, which maintains $k$ minimum-valued 
hashes of inserted elements, preserving Jaccard similarity at 
$O(k)$ comparison cost. Both produce fixed-size summaries of 
arbitrary-sized sets, which is 
what we need to represent 
neighborhoods of widely varying degree in a single-sized feature vector.

\vspace{-9pt}
\paragraph{Knowledge graph embeddings.}
KGE models learn a vector for each 
entity and each relation by optimizing a scoring function over 
$\mathcal{T}$. \textbf{TransE}~\cite{TransE} models relations 
as translations: $\mathbf{h} + \mathbf{r} \approx \mathbf{t}$. 
\textbf{DistMult}~\cite{distmult} uses a diagonal bilinear 
scoring that is symmetric in $\mathbf{h}$ and $\mathbf{t}$. 
\textbf{ComplEx}~\cite{complex} extends DistMult with 
complex-valued embeddings to capture asymmetric relations. 
\textbf{RotatE}~\cite{rotate} models relations as rotations in 
complex space and is widely used as a link-prediction decoder. 
KGE models train efficiently at scale and produce 
vectors that summarize each entity's relational role in the graph.

\vspace{-9pt}
\section{The \sysname{} Framework}
\label{sec:framework}
\vspace{-9pt}

Two design decisions distinguish the \sysname{} pipeline in Fig.~\ref{fig:model-arch} from prior preprocessing-based KG methods.
First, \sysname{} \emph{decomposes} structure into a local feature (the 1-hop neighborhood) and a global feature (the entity's position in the relational graph), because no single feature simultaneously captures both kinds of information at a fixed size.
Second, \sysname{} fuses these features with the learned embedding through a single scalar gate shared across all entities, which forces the model to learn a global balance between structural priors and learned refinement rather than per-entity weights.

For each entity $e \in \mathcal{E}$, \sysname{} computes three inputs: a local feature $f_L(e)$, a global feature $f_G(e)$, and a learned embedding $E(e)$ updated during task training. The two structural inputs $f_L$ and $f_G$ are computed once, in preprocessing, from the training triples $\mathcal{T}$ alone, a point we return to in Sec.~\ref{sec:fusion}. We describe the three options we try for $f_L$ (Sec.~\ref{sec:local}), the three options for $f_G$ (Sec.~\ref{sec:global}), and the fusion mechanism that combines them into a single representation $z(e)$ (Sec.~\ref{sec:fusion}). Sec.~\ref{sec:training} then describes how $z(e)$ enters training, including the choice of backbone and task-specific decoder.
\vspace{-9pt}
\subsection{Local Structural Prior: Neighborhood Encoding}
\label{sec:local}
\vspace{-9pt}

The local structural prior summarizes each entity's immediate 
relational context. Encoding the 1-hop neighborhood 
$\mathcal{N}(e)$ as a fixed-size vector requires choosing what 
information to preserve and what to compress, and \sysname{} 
considers three encodings that span this trade-off. The 
\emph{Bloom filter signature} hashes each $(r, x) \in 
\mathcal{N}(e)$ as the string ``$r\_x$'' into an $m$-bit 
array using $k$ hash functions, preserving the relation $r$ 
as well as the neighbor identity $x$. App.~\ref{app:bloom} 
gives the full construction procedure. \sysname{} defaults 
to the Bloom filter because it is simple and informative, 
but we also report results for MinHash and RelDist to show 
that the framework can accommodate other local features.
\vspace{-9pt}
\subsection{Global Structural Prior: Positional Encoding}
\label{sec:global}
\vspace{-9pt}

The global feature situates each entity in a continuous vector space whose geometry reflects the full graph's relational structure. \sysname{} computes this feature with a pre-trained KGE model: the chosen KGE is trained on the full set of training triples $\mathcal{T}$ before \sysname{}'s downstream task training begins, and its entity vectors are then frozen. 
Any KGE that produces fixed-size entity vectors fits the framework, but our default is TransE~\cite{TransE}
since it fits the 
global feature's purpose. 

The downstream model needs each entity's coordinate in a space that reflects
\begin{wrapfigure}{r}{0.5\linewidth}
  \centering
  \includegraphics[width=\linewidth]{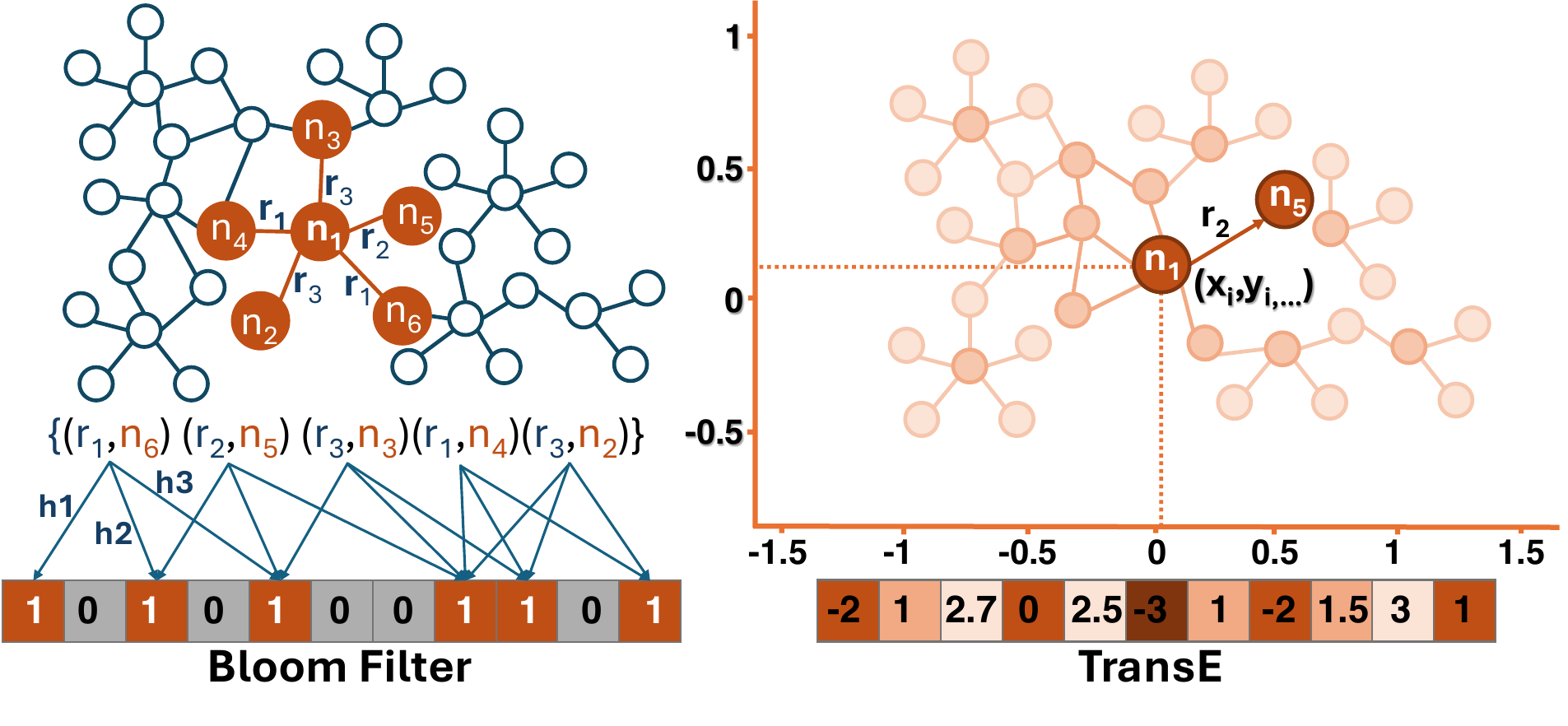}
  \caption{Local and global priors in \sysname{}. (Left) A Bloom filter
  hashes each $(r,n)$ pair of an entity's 1-hop neighborhood into $k$ bits.
  (Right) TransE places each entity at a point whose position reflects
  global relational structure.}
  \label{fig:bloom-transE}

\end{wrapfigure} the graph's relational structure as a whole, not a standalone scorer for triples. 

TransE's translational geometry fits this role naturally: each entity receives a single vector whose distance from other entities reflects relational proximity, and these distances are interpretable as inputs to the fusion mechanism.
More expressive KGEs, such as DistMult and ComplEx, model relations as relation-specific transformations (diagonal matrices and complex rotations, respectively), so their entity vectors are interpreted only in combination with those transformations. 
This makes them less natural as standalone position vectors, though we evaluate them as alternatives 
to test 
\sysname{}'s
robustness.

\vspace{-9pt}
\subsection{Gated Feature Fusion}
\label{sec:fusion}
\vspace{-8pt}

\sysname{} combines the local feature, global feature, learned embedding, and any domain-specific feature in two stages. 
In the first stage, each input feature is projected into a common $d$-dimensional latent space by a dedicated two-layer MLP with ReLU activation and dropout, then the projected vectors are concatenated and passed through a fusion MLP to produce a single structural representation:
\vspace*{-4pt}
\begin{equation}
h_{\text{struct}}(e) = \mathrm{MLP}_{\text{fuse}}\Big(
\mathrm{MLP}_L(f_L(e)) \,\big\|\,
\mathrm{MLP}_G(f_G(e)) \,\big\|\,
\mathrm{MLP}_X(x(e))\Big),
\end{equation}
where $\|$ denotes concatenation and $x(e)$ is an optional domain-specific feature.
The output $h_{\text{struct}}(e) \in \mathbb{R}^d$ has the same dimension as the learned embedding $E(e)$. 
In the second stage, the structural representation is combined with $E(e)$ through a single scalar gate $g \in [0, 1]$:
\begin{equation}
z(e) = g \cdot E(e) + (1-g) \cdot h_{\text{struct}}(e).
\end{equation}

The gate is learned end-to-end from task supervision and is shared across all entities. 
A per-entity gate, while more flexible, would let the model memorize a different balance for each entity, defeating the purpose of structural priors as a generalizing signal. 
A single shared scalar instead forces the model to learn one balance that applies uniformly across the graph.

During downstream training, $f_L(e)$, $f_G(e)$, and $x(e)$ remain frozen.
Only the learned embedding $E(e)$, the projection MLPs, the fusion MLP, the
gate, and the task-specific model parameters are updated. This separation
keeps the structural priors reusable across tasks and backbones, and it
bounds training-time memory by the active mini-batch rather than the full
graph.
\vspace{-9pt}
\paragraph{Data hygiene.}
Every structural prior is computed during preprocessing from the training triples $\mathcal{T}$ alone. The local signatures (Sec.~\ref{sec:local}) are built in a single pass over $\mathcal{T}$ (App.~\ref{app:bloom}), the global KGE (Sec.~\ref{sec:global}) is trained on $\mathcal{T}$ and then frozen. Negatives are drawn uniformly at random. No held-out triple is ever inserted, scored, or sampled. Validation triples serve only early stopping and model selection, and test triples only the final evaluation under the official splits. The two tasks differ only in what is held out: for link prediction it is a set of edges, none seen during feature construction; for node property prediction it is node labels, while the graph structure is shared across splits as it is for every baseline, and no label enters a structural feature. This boundary is what makes the reported gains attributable to generalization rather than leakage.
\vspace{-9pt}
\section{Training}
\label{sec:training}
\vspace{-9pt}

The fused representation $z(e)$ produced by \sysname{} is 
task-agnostic: the same representation drives both node property 
prediction and link prediction, and task-specific behavior is 
introduced only through the choice of backbone and decoder.

\vspace{-9pt}
\subsection{Computation and storage}
\vspace{-9pt}

The structural priors (Bloom filters, KGE vectors, domain-specific features) are stored as fixed-size tables in CPU memory and transferred to the GPU only for entities in the current mini-batch; no gradients flow to the frozen tables.  GPU memory is therefore bounded by the active mini-batch size, enabling training at MAG240M scale (244M entities).

\vspace{-9pt}
\subsection{Backbone GNN Configurations}
\label{sec:backbones}
\vspace{-9pt}

\emph{Relation-aware GNNs} (R-GCN, CompGCN) maintain 
per-relation parameters and model relational semantics with 
high fidelity. Their parameter count grows as 
$\mathcal{O}(|\mathcal{R}| d^2)$, which scales poorly when 
$|\mathcal{R}|$ is large, and \sysname{} reduces this burden by 
providing a structurally informative input representation that 
allows the GNN to reach strong accuracy with shallower or 
narrower architectures. \emph{Relation-agnostic and 
attention-based GNNs} (GraphSAGE, GAT, HGT) share parameters 
across relations and scale well, but discard edge-type 
information during message passing. \sysname{}'s global feature, 
derived from a relation-aware KGE, injects relational semantics 
into these architectures through the input representation, 
partly compensating for the lack of per-relation parameters. 
The \emph{decoder-only} configuration passes $z(e)$ directly to 
the task decoder with no GNN layers between fusion and decoder, 
so only the fusion parameters and the decoder are trained; this 
configuration avoids message-passing
entirely and tests 
whether the precomputed structural features carry sufficient 
signal on their own.

\vspace{-9pt}
\subsection{Decoders and Losses}
\label{sec:decoders}
\vspace{-9pt}

For node property prediction, a two-layer MLP classification 
head maps the entity representation to logits over the label 
set. Training minimizes cross-entropy loss over the labeled 
entities $\mathcal{E}_L$, and we report classification accuracy.

For link prediction, we use RotatE~\cite{rotate} as the scoring 
function: each relation is represented as a phase vector in 
complex space, and a triple is scored as the negative distance 
between the rotated head and the tail. Entity representations 
are split along the feature dimension into real and imaginary 
parts to obtain complex-valued inputs to RotatE. We train with the self-adversarial loss of~\cite{rotate}, where each negative is weighted by its difficulty score within the batch, with negatives drawn uniformly at random, and we report filtered MRR and Hits@k on both head and tail prediction.

\vspace{-7pt}
\section{Experiments}
\label{sec:experiments}
\vspace{-9pt}

\subsection{Experimental Setup}
\label{sec:exp-setup}
\vspace{-9pt}
Our experiments use Open Graph Benchmark datasets~\cite{OGB,OGBLSC},
specifically OGB-MAG and MAG240M for node property prediction and
OGB-WikiKG2 and OGB-BioKG for link prediction. We also use datasets from the
Heterogeneous Graph Benchmark~\cite{HGB} in Sec.~\ref{sec:exp-generalization}.
All experiments use the official train/validation/test splits provided by the
benchmarks. Hardware is a single NVIDIA L40S GPU (48\,GB) on a server with an
AMD Genoa CPU and 768\,GB RAM (App.~\ref{app:hardware}). Training uses AdamW
with learning rate $10^{-3}$. For all GNN backbones we use two layers, hidden
dimension 256, dropout 0.5, fanout $[25, 20]$ for node prediction; batch size
1024, margin $\gamma = 12$, $K = 32$ negatives, and self-adversarial sampling
for link prediction, except that R-GCN is trained full-batch on OGB-MAG and
with smaller settings on WikiKG2. Since each backbone uses identical
hyperparameters with and without \sysname{}, differences in performance can be
attributed solely to node features. \sysname{}'s Bloom filter and KGE
parameters are chosen per dataset (App.~\ref{app:structural-priors}), and full
training configurations are in App.~\ref{app:nodeprop-config}
and~\ref{app:lp-hyperparams}.

The experiments answer
four questions 
for both tasks: whether \sysname{}
improves accuracy across diverse GNN backbones
(Sections~\ref{sec:exp-node}--\ref{sec:exp-link}), whether the
framework is robust to the specific encoding chosen for each
prior (Sec.~\ref{sec:exp-robustness}), and whether the
benefit generalizes across schemas and domains
(Sec.~\ref{sec:exp-generalization}). End-to-end efficiency and resource usage are reported separately in Sec.~\ref{sec:exp-efficiency}.

\vspace{-4pt}
\subsection{Node Property Prediction}
\label{sec:exp-node}
\vspace{-9pt}

\begin{table*}[t]
\centering
\caption{Node property prediction accuracy (\%, mean $\pm$ std over five
seeds). \emph{Default} denotes the domain-specific features provided with
each benchmark: Word2Vec for OGB-MAG and RoBERTa for MAG240M. \emph{Local}
adds the Bloom filter. \emph{\sysname{}} adds both Bloom and TransE.
Bold: best accuracy on each dataset.}
\label{tab:node-default}
\small
\setlength{\tabcolsep}{5pt}
\begin{tabular}{l|ccc|ccc}
\toprule
& \multicolumn{3}{c|}{OGB-MAG (default = Word2Vec)}
& \multicolumn{3}{c}{MAG240M (default = RoBERTa)} \\
Model & Default &  Local &  \sysname{}
      & Default & Local &  \sysname{} \\
\midrule
R-GCN      & 37.67 $\pm$ 0.48 & 44.51 $\pm$ 0.48 & 47.98 $\pm$ 0.34
           & 68.00 $\pm$ 0.07 & 69.99 $\pm$ 0.05 & \textbf{75.04} $\pm$ 0.03 \\
GraphSAGE  & 31.66 $\pm$ 0.23 & 37.11 $\pm$ 0.34 & 43.37 $\pm$ 0.31
           & 66.83 $\pm$ 0.06 & 68.76 $\pm$ 0.09 & 71.87 $\pm$ 0.04 \\
GraphSAINT & 46.81 $\pm$ 0.48 & 50.26 $\pm$ 0.37 & \textbf{57.37} $\pm$ 0.26
           & 64.99 $\pm$ 0.23 & 68.30 $\pm$ 0.25 & 72.81 $\pm$ 0.08 \\
ClusterGCN & 36.03 $\pm$ 0.26 & 41.25 $\pm$ 0.53 & 45.56 $\pm$ 0.19
           & 61.13 $\pm$ 0.12 & 57.37 $\pm$ 0.11 & 72.58 $\pm$ 0.11 \\
GAT        & 31.87 $\pm$ 0.49 & 39.36 $\pm$ 0.28 & 49.10 $\pm$ 0.29
           & 66.82 $\pm$ 0.04 & 70.09 $\pm$ 0.09 & 71.95 $\pm$ 0.05 \\
HGT        & 43.43 $\pm$ 0.52 & 44.79 $\pm$ 0.61 & 53.39 $\pm$ 0.65
           & 57.76 $\pm$ 0.06 & 65.67 $\pm$ 0.07 & 74.52 $\pm$ 0.11  \\
\bottomrule
\end{tabular}
\vspace{-9pt}
\end{table*}

\begin{wraptable}{r}{0.50\linewidth}
\vspace{-12pt}
\centering
\caption{Micro-\,/\,Macro-F1 (\%, mean $\pm$ std over five seeds) for
\sysname{} on three heterogeneous graph benchmarks with GraphSAGE. SOTA:
PHGT as reported in~\cite{PHGT}. Bold: best per row.}
\label{tab:hgb-results}
\small
\setlength{\tabcolsep}{2pt}
\begin{tabular}{l|ccc}
\toprule
& w/o \sysname{} & \sysname{} & SOTA \\
\midrule
\multicolumn{4}{l}{\emph{DBLP}} \\
\quad Micro-F1 & 82.50 $\pm$ 0.31 & \textbf{97.52} $\pm$ 0.66 & 95.33 \\
\quad Macro-F1 & 81.70 $\pm$ 0.36 & \textbf{97.04} $\pm$ 0.79 & 94.96 \\
\midrule
\multicolumn{4}{l}{\emph{ACM}} \\
\quad Micro-F1 & 91.08 $\pm$ 0.34 & \textbf{95.56} $\pm$ 0.74 & 93.72 \\
\quad Macro-F1 & 91.20 $\pm$ 0.39 & 93.16 $\pm$ 0.70 & \textbf{93.79} \\
\midrule
\multicolumn{4}{l}{\emph{IMDb}} \\
\quad Micro-F1 & 65.28 $\pm$ 0.76 & \textbf{69.14} $\pm$ 1.20 & 68.81 \\
\quad Macro-F1 & 61.89 $\pm$ 0.88 & \textbf{66.14} $\pm$ 1.49 & 65.91 \\
\bottomrule
\end{tabular}
\vspace{-9pt}
\end{wraptable}

Node property prediction assigns a label to each entity using graph
structure, labeled neighbors, and any available domain-specific features.
We evaluate \sysname{} across six GNN backbones spanning relation-aware,
relation-agnostic, sampling-based, and attention-based paradigms (R-GCN,
GraphSAGE, GraphSAINT, ClusterGCN, GAT, HGT) on two variants of the
Microsoft Academic Graph, which represents bibliographic information:
\textbf{OGB-MAG} ($\sim$1.9M nodes, predicting venue across 349 classes) and
\textbf{MAG240M} ($\sim$244M nodes, predicting one of 153 subject areas for 1.4M
arXiv papers). The two datasets differ in two ways that matter for this
experiment: MAG240M is $126\times$ larger than OGB-MAG, and each dataset
comes with its own domain-specific features, namely, text embeddings of
each paper's title and abstract ($128$-dimensional Word2Vec~\cite{word2vec}
for OGB-MAG and $768$-dimensional RoBERTa~\cite{roberta} for MAG240M).
Word2Vec and RoBERTa both summarize text, but RoBERTa is more powerful and
more expensive to compute, store, and feed through the first GNN layer. We
refer to these as \emph{default features}.

The most direct test of \sysname{} is whether it improves a backbone
trained with the dataset's own default features.
Tab.~\ref{tab:node-default} reports test accuracy for each backbone under
three conditions: \emph{Default} (the dataset's default features alone),
\emph{Local} (default features plus the Bloom filter), and
\emph{\sysname{}} (default features plus both Bloom and TransE). The
results are consistent across datasets and backbones: adding \sysname{} on
top of the default features always improves accuracy, by far more than the
seed standard deviation. On OGB-MAG, the gain ranges from $+9.5$
percentage points on ClusterGCN to $+17.2$ on GAT. On MAG240M, it ranges
from $+5.0$ on GraphSAGE to $+16.8$ on HGT. The gains hold across
relation-aware, relation-agnostic, sampling-based, and attention-based
architectures, so the benefit is not tied to any particular paradigm. The
best accuracy is achieved by GraphSAINT + \sysname{} on OGB-MAG
($57.37\%$) and by R-GCN + \sysname{} on MAG240M ($75.04\%$). The
\emph{Local} column represents an ablation showing that adding the Bloom
filter alone, without TransE, achieves a substantial portion of the full
\sysname{} gain on most backbones, indicating that the local neighborhood
signal carries information the default features do not. The exception is
ClusterGCN on MAG240M, where the Bloom filter alone lowers accuracy and the
global prior is needed.

Another interesting test of \sysname{} is whether it can \emph{replace}
the default features not just complement them. The default features are
informative but expensive, and we show in App.~\ref{app:nodeprop-no-text}
that training with \sysname{} on the graph structure only without default
features outperforms training with the default features but without
\sysname{}.

\vspace{-9pt}
\subsubsection{Generalization Across Heterogeneous Domains}
\label{sec:exp-generalization}
\vspace{-4pt}

To test whether \sysname{} generalizes beyond the OGB datasets, we
evaluate on three node classification tasks on heterogeneous graph
datasets from HGB~\cite{HGB}, spanning different schemas and domains: DBLP
and ACM (academic bibliography) and IMDb (movie-genre prediction).
Tab.~\ref{tab:hgb-results} reports Micro-F1 and Macro-F1 for GraphSAGE,
the simplest of our backbones, with and without \sysname{} (Bloom +
TransE). \sysname{} improves both metrics on every dataset, with the
largest gains on DBLP (+15.0 Micro-F1) where citation neighborhood
structure is highly discriminative.

These numbers position \sysname{} favorably against the recent SOTA
PHGT~\cite{PHGT} (also shown in the table), a heterogeneous graph
transformer specifically engineered for these benchmarks. The contribution
here is threefold: \emph{generality}, since the gains hold across both
academic bibliography and entertainment domains; \emph{efficiency}, since
the same \sysname{} recipe works on every benchmark with no
dataset-specific tuning, while PHGT uses a heavyweight transformer trained
per-dataset; and \emph{strength of priors}, since precomputed structural
features fed into a simple backbone match or exceed SOTA accuracy on
graphs whose leaderboards are otherwise dominated by purpose-built
architectures.

\vspace{-9pt}

\subsection{Link Prediction}
\label{sec:exp-link}
\vspace{-9pt}
Link prediction ranks candidate entities for a specific $(h, r, ?)$ or $(?, r, t)$ query.
We evaluate on two OGB benchmarks~\cite{OGB} that span a $25\times$ range
in scale: \textbf{OGB-WikiKG2} ($2.5$M entities,
$535$ relations, $16.1$M training triples, $500$ sampled
negatives per query under filtered MRR), and
\textbf{OGB-BioKG} ($94$K entities across five
biomedical types, with type-restricted candidates that eliminate
trivial cross-type distinctions). WikiKG2 stresses the long-tail
behavior of structural priors at scale: over $80\%$ of its
entities appear in fewer than 10 training triples, and the
long tail concentrates on the head side because tail entities
(e.g., countries, organizations) tend to be shared across many heads.
More detailed dataset-level statistics appear in
App.~\ref{app:datasets}.

\vspace{-4pt}
\subsubsection{WikiKG2: Structural Priors Across Architectures}
\label{sec:exp-link-headline}
\vspace{-4pt}
\begin{table}[t]
\centering
\small
\setlength{\tabcolsep}{2pt}

\vspace{-9pt}

\begin{minipage}[t]{0.55\linewidth}
\centering
\setlength{\tabcolsep}{3pt}
\caption{WikiKG2 link prediction (mean $\pm$ std over five seeds).
$^\ast$ indicates results reported in \citep{nodepiece,astartnet}.}
\label{tab:wikikg2-main}
\vspace{1pt}
\begin{tabular}{l|cc}
\toprule
Method & Val.\ MRR & Test MRR \\
\midrule
NodePiece + AutoSF$^\ast$ & 0.5806 & 0.5703 \\
NBFNet & OOM & OOM \\
A\textsuperscript{$\star$}Net$^\ast$ & 0.6851 & 0.6767 \\
\midrule
RotatE & 0.4352\std{0.0004} & 0.4341\std{0.0009} \\
\textbf{RotatE + \sysname{}} & \textbf{0.7032}\std{0.0014} & \textbf{0.6941}\std{0.0016} \\
\midrule
GraphSAGE & 0.5720\std{0.0060} & 0.5689\std{0.0040} \\
GraphSAGE + \sysname{} & 0.6216\std{0.0061} & 0.6129\std{0.0024} \\
R-GCN & 0.3717\std{0.0233} & 0.3851\std{0.0225} \\
R-GCN + \sysname{} & 0.5953\std{0.0031} & 0.5957\std{0.0039} \\
CompGCN & 0.6275\std{0.0033} & 0.6256\std{0.0039} \\
CompGCN + \sysname{} & 0.6661\std{0.0016} & 0.6562\std{0.0010} \\
\bottomrule
\end{tabular}
\end{minipage}\hfill
\begin{minipage}[t]{0.45\linewidth}
\centering
\vspace{0pt}
\begin{minipage}{0.85\linewidth}
\centering
\includegraphics[width=\linewidth]{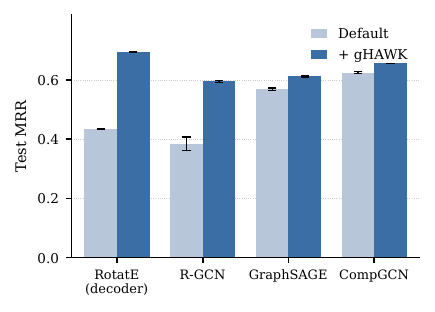}
\captionof{figure}{Test MRR on WikiKG2 with and without \sysname{}
(mean $\pm$ std over five seeds). Gains are largest for RotatE (+26.0) and
R-GCN (+21.1), which capture the least structure on their own. }
\label{fig:lp-gains}
\end{minipage}
\end{minipage}

\vspace{-9pt}

\begin{minipage}[t]{0.49\linewidth}
\centering
\caption{Head MRR on WikiKG2 (decoder-only) stratified by training frequency. Single run result.}
\label{tab:head-mrr-frequency}

\begin{tabular}{r|rccc}
\toprule
Train.\ triples & \% queries & RotatE & \sysname{} &  $\Delta$ \\
\midrule
1--5      & 20.12  & 0.0109 & \textbf{0.3497} & $+0.3388$ \\
6--20     & 71.77  & 0.0714 & \textbf{0.4658} & $+0.3944$ \\
21--100   & 7.17   & 0.3747 & \textbf{0.7147} & $+0.3400$ \\
101--1000 & 0.72   & 0.6355 & \textbf{0.7272} & $+0.0917$ \\
1000$+$   & 0.21   & \textbf{0.8989} & 0.7380 & $-0.1609$ \\
\midrule
All       & 100.00 & 0.0868 & \textbf{0.4627} & $+0.3759$ \\
\bottomrule
\end{tabular}
\end{minipage}\hfill
\begin{minipage}[t]{0.49\linewidth}
\centering
\caption{Head MRR on WikiKG2 (R-GCN) stratified by training frequency.  Single run result.}
\label{tab:head-mrr-frequency-rgcn}
\vspace{2pt}
\begin{tabular}{r|rccc}
\toprule
Train.\ triples & \% queries & R-GCN & \sysname{} &  $\Delta$ \\
\midrule
1--5      & 20.12  & 0.3852 & \textbf{0.5432} & $+0.1580$ \\
6--20     & 71.77  & 0.4822 & \textbf{0.5762} & $+0.0940$ \\
21--100   & 7.17   & 0.5091 & \textbf{0.5805} & $+0.0713$ \\
101--1000 & 0.72   & 0.4598 & \textbf{0.5513} & $+0.0915$ \\
1000$+$   & 0.21   & 0.4300 & \textbf{0.5295} & $+0.0995$ \\
\midrule
All       & 100.00 & 0.4644 & \textbf{0.5696} & $+0.1052$ \\
\bottomrule
\end{tabular}
\end{minipage}
\vspace{-4pt}
\end{table}
Tab.~\ref{tab:wikikg2-main} reports MRR on WikiKG2 across four
configurations: a RotatE decoder fed only \sysname{}'s
structural features (no message passing), and three GNN
backbones trained with and without \sysname{} features. The
backbones span the heterogeneous-GNN design space: R-GCN is
fully relation-aware, CompGCN
treats relations as compositional operators, and GraphSAGE
shares parameters across all relations. 
For competitive context, we compare with the reported results of
NodePiece + AutoSF~\cite{nodepiece}, NBFNet~\cite{nbfnet} (which exhausts GPU
memory, as also reported in~\cite{astartnet}), and A\textsuperscript{$\star$}Net~\cite{astartnet}.

In the RotatE decoder-only pipeline,  \sysname{}  improves test MRR from
$0.4341$ to $0.6941$, a $+26$ percentage point gain
achieved without any message passing. Adding \sysname{} features
to message-passing backbones also improves accuracy in every case: R-GCN gains $+21.1$ points, GraphSAGE gains $+4.4$ points, and CompGCN $+3.1$. Across architectures, structural priors provide a
consistent benefit.
With structural priors, 
even a
decoder-only architecture like RotatE with no aggregation matches or exceeds full
message-passing alternatives, supporting our claim that on
large KGs, the quality of input representations matters more
than the depth of aggregation.

The mechanism behind these gains is that \sysname{}
provides the backbone with an accurate representation of the long-tail of the entity-relation distribution.
In any embedding-based model, learned vectors only carry signal
where training provides enough gradient updates to move the
embedding away from initialization.
Since a large fraction of
WikiKG2 entities appear in fewer than five training triples, their $250$-dimensional embeddings receive too few updates and are essentially uninformative.
If we separate the test set into queries where the head is masked $(?, r, t)$ and ones where the tail is masked $(h, r, ?)$, we find that uninformative representations are starkest on the head side.
Tab.~\ref{tab:head-mrr-frequency} quantifies this.
The table groups the head entities into buckets based on how many training triples they appear in.
For each bucket, the table shows the fraction of queries for heads in this bucket, the MRR achieved by RotatE alone, the MRR achieved by RotatE + \sysname{}, and the gain that \sysname{} adds.
The first two rows show that RotatE alone has very low MRR for heads that appear in few training triples (20 or fewer).
\sysname{} adds a gain of over $+30$ percentage points in these cases.
These two rows are noteworthy since they account for more than $90\%$ of the queries.
The gain shrinks above $100$ training triples and turns negative in the 1000+ bucket.
Comparing Tab.~\ref{tab:wikikg2-main} and~\ref{tab:head-mrr-frequency}, we see that RotatE has low accuracy for head queries (near-random against $500$ negatives), and the $0.4341$ test MRR of RotatE comes almost entirely from the tail side. 
\sysname{} has good accuracy for head queries and overall.
A
Bloom filter deterministically encodes each entity's
neighborhood from the training graph regardless of training
frequency, and a pre-trained global embedding propagates signal
through shared relations and neighbors even to rare entities.
Structural features restore predictive capacity exactly where
learned embeddings have none.
\vspace{-4pt}
\subsubsection{Generalization to Biomedical Link Prediction}
\label{sec:exp-link-biokg}
\vspace{-4pt}
\begin{wraptable}{r}{0.50\linewidth}
\vspace{-6pt}
\centering
\caption{MRR on OGB-BioKG. Top rows: aggregate validation and test performance. Bottom rows: per-relation breakdown (test).}
\label{tab:biokg-results}
\small
\setlength{\tabcolsep}{4pt}
\begin{tabular}{l|cc|c}
\toprule
 & RotatE & \textbf{\sysname{}} & $\Delta$ \\
\midrule
\multicolumn{4}{l}{\textit{Aggregate}} \\
\quad Validation  & 0.7997 & \textbf{0.8380} & +0.038 \\
\quad Test   & 0.7989 & \textbf{0.8378} & +0.039 \\
\midrule
\multicolumn{4}{l}{\textit{Per-relation (test)}} \\
\quad drug $\to$ disease      & 0.531 & \textbf{0.628} & +0.096 \\
\quad drug $\to$ drug         & 0.617 & \textbf{0.704} & +0.088 \\
\quad disease $\to$ protein   & 0.587 & \textbf{0.636} & +0.049 \\
\quad drug $\to$ side-effect  & 0.204 & \textbf{0.239} & +0.035 \\
\quad drug $\to$ protein      & 0.804 & \textbf{0.832} & +0.028 \\
\quad protein $\to$ function  & 0.844 & \textbf{0.867} & +0.023 \\
\quad function $\to$ function & 0.942 & \textbf{0.962} & +0.020 \\
\quad protein $\to$ protein   & 0.969 & \textbf{0.973} & +0.004 \\
\bottomrule
\end{tabular}
\vspace{-9pt}
\end{wraptable}

We test whether \sysname{}'s benefits generalize beyond
WikiKG2 by evaluating on OGB-BioKG, a biomedical knowledge
graph that differs from WikiKG2 in three ways. First, BioKG
uses \emph{type-restricted evaluation}: when ranking the tail
of a query like (Drug X, treats, ?), candidates are restricted
to entities of the correct type (Disease) rather than the full
entity set, eliminating trivial cross-type distinctions.
Second, BioKG has a denser degree distribution (median degree
$22$) so most entities are well-represented in training.
Third, its dominant relations
(drug $\to$ drug, protein $\to$ protein) are symmetric, unlike most KGs
where asymmetry is the norm. We therefore use ComplEx as the KGE for the 
global feature for BioKG, 
since it 
naturally encodes symmetric and asymmetric relations alike.

\sysname{} improves test MRR from $0.7989$ to $0.8378$, a
$+3.9$-point gain that closes nearly $20\%$ of the headroom
remaining above RotatE. The per-relation breakdown in
Tab.~\ref{tab:biokg-results} shows that this gain concentrates
on the clinically important predictions where the base model
has the lowest accuracy. 
Large gains appear on drug $\rightarrow$ disease, drug $\rightarrow$ drug, and drug $\rightarrow$ side-effect.
This mirrors what we observed on WikiKG2, where \sysname{}'s
gain similarly concentrated on entities the base model
handled poorly. 
\vspace{-9pt}
\subsection{Robustness to the Choice of Prior}
\label{sec:exp-robustness}
\vspace{-8pt}

\begin{table}[t]
\vspace{-10pt}
\centering
\small
\caption{Robustness of \sysname{} to the choice of local/global prior.
Left: node property prediction on OGB-MAG (test
accuracy \%, GraphSAINT, default Word2Vec features). Right: link
prediction on WikiKG2 (MRR, RotatE).  Single run result.
}
\label{tab:robustness-both}
\setlength{\tabcolsep}{3pt}
\begin{minipage}[t]{0.40\linewidth}
\centering
\subcaption{Node property prediction.}
\label{tab:robustness-node}
\begin{tabular}{l|ccc|}
\toprule
Local $\downarrow$ / Global $\rightarrow$ & TransE & DistMult & ComplEx \\
\midrule
Bloom    & \textbf{57.97} & 57.14 & 57.38 \\
RelDist  & 56.65 & 56.36 & 56.59 \\
MinHash  & 56.06 & 55.41 & 55.33 \\
\bottomrule
\end{tabular}
\end{minipage}\hfill
\begin{minipage}[t]{0.54\linewidth}
\centering
\subcaption{Link prediction (Test MRR / Head MRR).}
\label{tab:robustness-link}
\begin{tabular}{l|ccc}
\toprule
Local $\downarrow$ / Global $\rightarrow$ & TransE & DistMult & ComplEx \\
\midrule
Bloom    & \textbf{.693}/.463 & \textbf{.693}/.463 & .692/.462 \\
RelDist  & .686/.458 & .677/.451 & .681/.461 \\
MinHash  & .680/.453 & .675/.439 & .685/.453 \\
\bottomrule
\vspace{-9pt}
\end{tabular}
\vspace{-9pt}
\end{minipage}
\vspace{-9pt}
\end{table}

\sysname{} treats Bloom + TransE as the default
local/global prior combination, but the framework itself 
can be used with other choices.
To verify this, 
we evaluate combinations of three local
priors (Bloom, RelDist, MinHash) and three global priors
(TransE, DistMult, ComplEx) on both tasks,
using the
strongest backbone for each: GraphSAINT on OGB-MAG for node
property prediction 
and the RotatE
decoder-only configuration on WikiKG2 for link prediction.

The two tasks tell the same story. 
On
node property prediction (Tab.~\ref{tab:robustness-node}), the spread
across the entire grid is under three percentage points, and the default
Bloom + TransE cell has the best accuracy.
Even the weakest combination (MinHash +
ComplEx at $55.33\%$) remains well above the Default
($46.81\%$) and Default + Local ($50.26\%$) baselines from
Tab.~\ref{tab:node-default}. On link prediction
(Tab.~\ref{tab:robustness-link}), every combination converges
to an overall MRR of about $0.69$ and a head MRR of about $0.46$, with the default Bloom + TransE having the highest MRR.
Across both tasks, the benefit of \sysname{} comes from the presence of both a local and a global structural prior, not from the specific encoding chosen (App.~\ref{app:lp-single-prior} isolates the contribution of each prior), and we adopt Bloom + TransE as the default because it is the simplest to construct, the cheapest to store, and one of the most efficient to pre-train.
The full costs of preprocessing for different priors appear in
App.~\ref{app:preprocessing-costs}.

We observe that every choice of structural prior lifts head MRR from $0.087$
for RotatE alone (Tab.~\ref{tab:head-mrr-frequency}) to about $0.46$
(Tab.~\ref{tab:robustness-link}), because the priors represent the long-tail
of the head-entity distribution. Head MRR never exceeds this level, however,
even though ample headroom remains, and the plateau is not the result of a
weak choice of prior. Proposition~A (App.~\ref{app:proofs}) shows why: any
query-independent feature, whether hashed (Bloom), counted (RelDist),
pre-trained (TransE, DistMult, ComplEx), or jointly learned, produces
representations of two entities that grow arbitrarily close as their training
neighborhoods converge (for trained features, under the local assumptions of
App.~\ref{app:proofs}), and so do their decoder scores. Empirically, $93\%$ of
\sysname{}'s rank-$1$ wrong predictions are \emph{doppelgangers} of the
correct answer, and an oracle that resolves doppelganger confusion
(App.~\ref{app:lp-oracle}) attains $0.934$ head MRR. The gap between the
plateau and the oracle is therefore the cost of doppelganger confusion, and
closing it requires conditioning on the query (App.~\ref{app:wl}).

\vspace{-9pt}
\subsection{Efficiency and Resource Usage}
\label{sec:exp-efficiency}
\vspace{-5pt}

Because the priors are frozen and transferred to GPU by mini-batch
(Sec.~\ref{sec:training}), the cost \sysname{} adds is modest. Preprocessing
is a one-time expense shared by every backbone, seed, and ablation (about
25\,min on OGB-MAG, 2\,h on WikiKG2, and 6.5\,h on MAG240M;
App.~\ref{app:preprocessing-costs}). On a GNN backbone the fusion module
adds under $0.1\%$ of parameters. It changes total training time on OGB-MAG
by between $-9\%$ and $+19\%$, and raises per-epoch time on WikiKG2 by
5--46\%. The decoder-only pipeline, where the priors replace the backbone,
costs $+22\%$ per epoch for its $+0.26$ MRR gain. Peak GPU memory is
dominated by the backbone and entity table rather than the priors
($+4.6\%$ on WikiKG2), which is what allows training on MAG240M's $244$M
entities on a single GPU. Per-backbone measurements are in
App.~\ref{app:efficiency}. 
\vspace{-9pt}
\section{Related Work}
\label{sec:related-work}
\vspace{-9pt}

\sysname{} sits at the intersection of six lines of work.
Sampling-based GNNs (GraphSAGE~\cite{graphSAGE},
FastGCN~\cite{fastgcn}, ClusterGCN~\cite{clustergcn},
GraphSAINT~\cite{graphsaint}) make large-graph training feasible by restricting each step to a graph subset; \sysname{}
is \emph{complementary}, shifting structural capture to preprocessing so that Bloom and TransE features remain fully
informative regardless of sampling. Precomputed structural features (DeepWalk~\cite{deepwalk}, node2vec~\cite{node2vec},
NodePiece~\cite{nodepiece}, SIGN~\cite{sign}, Bloom signatures~\cite{bloomsig})
augment GNN inputs with fixed-size summaries; \sysname{}
extends this to heterogeneous KGs by combining a frozen local
prior with a global KGE through a gated fusion that operates
on top of the learned embeddings. Methods that combine local and global information in other domains~\cite{ding2025global, du2024hierarchical, eliasof2024global, chen2025local, xu2021self} learn both representations jointly with the task, whereas \sysname{}'s priors are computed once and reused across tasks and backbones (App.~\ref{app:related-extended}). Heterogeneous graph
transformers (HGT~\cite{hgt}, PHGT~\cite{PHGT}) are
purpose-built for multi-type graphs and achieve SOTA on
benchmarks like DBLP and IMDb through dataset-specific
architectures; \sysname{} matches or exceeds them with a
simple GraphSAGE backbone and a single Bloom + TransE recipe
applied without per-dataset tuning
(Sec.~\ref{sec:exp-generalization}). Query-conditioned methods (NBFNet~\cite{nbfnet},
A\textsuperscript{$\star$}Net~\cite{astartnet}) compute query-specific
representations to escape the limit on any 
query-independent feature, but at a per-query cost: at WikiKG2
scale, NBFNet exhausts memory and A\textsuperscript{$\star$}Net's
accuracy falls below \sysname{}'s decoder-only configuration
(Sec.~\ref{sec:exp-link-headline});
\sysname{} trades this expressiveness for the scalability of
precomputed query-independent features. Long-tail
analysis~\cite{popularity} documents that embedding models
degrade for rare entities, and existing approaches address
this indirectly through training-time interventions such as
adjusted negative sampling or loss reweighting; \sysname{}
addresses it directly through features that remain
informative regardless of the frequency of training triples.
\vspace{-9pt}

\section{Limitations and Future Work}
\vspace{-9pt}
\label{sec:limitations}

Three design choices give \sysname{} its scalability and generality: its priors are query-independent, they are combined with the learned embedding through a single gate shared across all entities, and they are computed once, in preprocessing, over a static training graph. Each choice also bounds what the method can do, and we discuss the three in turn.

Because the priors are query-independent, they cannot separate doppelganger entities (Proposition~A, App.~\ref{app:proofs}), which caps head MRR on WikiKG2 well below what an oracle reaches (Sec.~\ref{sec:exp-robustness}). The same limit applies to message-passing GNNs bounded by 1-WL (App.~\ref{app:wl}); for gradient-trained features, the bound assumes local strong convexity around the trained solution (Lemma 3), which KGE objectives do not guarantee globally. The ceiling is the link-prediction form of the local isomorphism problem, which labeling tricks~\cite{seal2} resolve by conditioning the representation on the query. Applied to every candidate, however, query conditioning reintroduces the per-query cost that \sysname{} is built to avoid (Sec.~\ref{sec:related-work}). Our error analysis points to a cheaper route: $93\%$ of \sysname{}'s rank-1 errors are doppelgangers of the correct answer (App.~\ref{app:lp-oracle}), so the errors that remain arise among near-identical candidates competing for the top rank. This motivates query-aware reranking as a general direction.

The shared gate (Sec.~\ref{sec:fusion}) learns one balance between structural priors and learned embeddings for the whole graph. That balance suits the long tail, where most queries lie and learned embeddings are weakest, but it cannot adapt to the few entities whose embeddings are already strong. For the $0.21\%$ of WikiKG2 queries whose head appears in more than $1000$ training triples, RotatE + \sysname{} falls below RotatE alone ($0.738$ versus $0.899$ head MRR, Tab.~\ref{tab:head-mrr-frequency}). This regression does not occur with R-GCN, whose head MRR in this bucket is only $0.430$: with no strong embedding for the priors to dilute, R-GCN + \sysname{} improves on R-GCN in every 
bucket (Tab.~\ref{tab:head-mrr-frequency-rgcn}). A degree-conditioned gate, whose balance varies with an entity's training frequency, is a possible solution for the RotatE case.

Computing the priors once, in preprocessing, adds a one-time cost amortized across every backbone, seed, and ablation (Sec.~\ref{sec:exp-efficiency}), and requires that preprocessing examine only training triples, as ours does (Sec.~\ref{sec:fusion}); a pipeline adapted to a new dataset that admitted held-out edges would silently inflate link-prediction results. It also ties \sysname{} to static-graph training. The local prior updates inductively, with two $\mathcal{O}(k)$ Bloom insertions per new edge, so an unseen entity receives its local prior as soon as its edges appear, while the global KGE prior is refreshed periodically, at a cost of about two hours on WikiKG2. Evaluating \sysname{} under genuine graph dynamics, with an inductive KGE as the global prior, is left to future work.

\vspace{-9pt}
\section{Conclusion}
\vspace{-9pt}
We presented \sysname{}, a framework for scalable GNN training on KGs that shifts the burden of capturing graph structure from iterative message passing to a one-time preprocessing step. Each entity is represented by a local feature (a Bloom filter encoding its neighborhood) and a global feature (a pre-trained TransE embedding), fused with the learned embedding through a gated mechanism. Our experiments show that these structural priors provide substantial benefit. For node property prediction, \sysname{} consistently improves accuracy across six GNN backbones, spanning relation-aware, relation-agnostic, sampling-based, and attention-based architectures. For link prediction, \sysname{} enables a RotatE decoder-only pipeline with no message passing to improve test MRR on WikiKG2 from $0.43$ to $0.69$, a $26$ percentage point gain, and to outperform message-passing GNNs. This gain comes from accurately representing the long-tail of the head entity-relation distribution. The same effect transfers to BioKG, where \sysname{} improves most on the clinically important drug relations. Beyond these results, we identified a structural property of
query-independent representations, which we call doppelganger confusion:
entities whose training neighborhoods nearly coincide receive nearly identical
representations, and therefore nearly identical scores, whether the features
are hashed, counted, or learned (Proposition~A). This creates a ceiling that
no choice of local or global prior escapes in our experiments: head MRR on
WikiKG2 stays at or below $0.46$ for all combinations of priors that we tested, while an oracle
that resolves the confusion reaches $0.93$. This characterization delineates
the scope of the structural-prior paradigm, separating the regime where better
encodings improve predictions from the regime that requires query
conditioning. Such query conditioning is a natural extension of \sysname{}. More broadly, our results
suggest that on large KGs, the quality of input representations matters more
than the depth of aggregation.

\begin{ack}
This work was funded by the University of Texas at Arlington, the University of Texas System through a Science and Technology Acquisition and Retention (STARs) grant, and the Texas Advanced Computing Center (TACC) through a computing allocation.
Humera Sabir acknowledges support from a Fulbright Program grant
sponsored by the Bureau of Educational and Cultural Affairs of
the U.S. Department of State and the United States Educational
Foundation in Pakistan (USEFP), and administered by the
Institute of International Education (IIE).

The authors declare no competing interests.
\end{ack}

\bibliographystyle{abbrvnat}

\bibliography{gnn_neurips}
\clearpage
\appendix

\section*{Appendix: Table of Contents}
\addcontentsline{toc}{section}{Appendix}
\vspace{0.5em}

\begin{itemize}[label={}, leftmargin=0pt, itemsep=2pt]
  \item \ref{app:datasets}.~\textbf{Hardware and Dataset Statistics} \dotfill \pageref{app:datasets}
  \item \ref{app:structural-priors}.~\textbf{Structural Priors: Implementation and Preprocessing Costs} \dotfill \pageref{app:structural-priors}
  \item \ref{app:efficiency}.~\textbf{Training and Inference Cost} \dotfill \pageref{app:efficiency}
  \item \ref{app:nodeprop}.~\textbf{Node Property Prediction: Extended Results} \dotfill \pageref{app:nodeprop}
  \item \ref{app:linkpred}.~\textbf{Link Prediction: Extended Results} \dotfill \pageref{app:linkpred}
  \item \ref{app:proofs}.~\textbf{Theoretical Analysis of the Doppelganger Effect} \dotfill \pageref{app:proofs}
  \item \ref{app:related-extended}.~\textbf{Extended Related Work} \dotfill \pageref{app:related-extended}
\end{itemize}

\vspace{1em}
\noindent\rule{\linewidth}{0.4pt}

\section{Hardware and Dataset Statistics}
\label{app:datasets}

\subsection{Hardware}
\label{app:hardware}
All experiments used a single server with an AMD Genoa CPU,
768\,GB RAM, and four NVIDIA L40S GPUs (48\,GB each); each run
used one GPU.

\subsection{Node Property Prediction Datasets}

Tab.~\ref{tab:app-node-datasets} reports statistics for both
OGB and HGB node property prediction benchmarks.

\begin{table}[ht]
\centering
\caption{Node property prediction dataset statistics.}
\label{tab:app-node-datasets}
\setlength{\tabcolsep}{4pt}
\begin{tabular}{l|rrrrl}
\toprule
Dataset & Nodes & Edges & Entity types & Rel.\ types & Text feature (\emph{Default}) \\
\midrule
OGB-MAG  & 1.94M & 21M   & 4 & 4 & Word2Vec (128-d) \\
MAG240M  & 244M  & 1.3B  & 3 & 3 & RoBERTa (768-d) \\
\midrule
DBLP     & 26K   & 239K  & 4 & 6 & BoW (334-d) \\
ACM      & 10K   & 547K  & 4 & 8 & BoW (1902-d) \\
IMDb     & 21K   & 86K   & 4 & 6 & Metadata vector (3489-d) \\
\bottomrule
\end{tabular}
\end{table}

OGB-MAG and MAG240M represent the Microsoft Academic Graph at
different scales.  The task is paper venue prediction on OGB-MAG
(349 classes) and paper subject area prediction on MAG240M
(153 classes).  Word2Vec features encode paper titles. RoBERTa
features encode paper abstracts.  DBLP, ACM, and IMDb are drawn
from the Heterogeneous Graph Benchmark~\cite{HGB} and serve as
generalization targets outside the OGB family.
\subsection{Link Prediction Datasets}

Tab.~\ref{tab:app-lp-datasets} reports statistics and 
evaluation protocols for both link prediction benchmarks.

\begin{table}[ht]
\centering
\caption{Link prediction dataset statistics.}
\label{tab:app-lp-datasets}
\setlength{\tabcolsep}{3pt}
\begin{tabular}{l|rrrrl}
\toprule
Dataset & Entities & Relations & Train triples & Eval.\ neg. & Protocol \\
\midrule
WikiKG2 & 2{,}500{,}604 & 535 & 16{,}109{,}182 & 500 & filtered MRR \\
BioKG   & 93{,}773      &  51 & 4{,}762{,}677  & 500 & same-type filtered MRR \\
\bottomrule
\end{tabular}
\end{table}

\paragraph{WikiKG2.}
WikiKG2 is a large general-purpose knowledge graph derived 
from Wikidata. Its entities span people, places, 
organizations, and abstract concepts, with $535$ relation 
types covering biographical, geographical, and categorical 
predicates. Each test query is evaluated against $500$ 
sampled candidate entities under the standard filtered MRR 
protocol, where triples present in the training, validation, 
or test set are removed from the candidate ranking. The 
dataset's degree distribution is heavily skewed: over $80\%$ 
of entities appear in fewer than 10 training triples, with 
the long tail concentrated on the head side because tail 
entities (e.g., countries, organizations, common categories) tend 
to be shared across many heads.

\paragraph{BioKG.}
BioKG is a biomedical knowledge graph organized around five 
entity categories: drugs, diseases, proteins, functions, and 
side-effects. The graph is denser than WikiKG2 in absolute 
terms, with a median entity degree of $22$, and most 
entities are well-represented in training. BioKG uses 
\emph{type-restricted evaluation}: candidates for a query 
$(?, r, t)$ are restricted to entities of the type expected 
by relation $r$. For example, candidates for the head of 
(?, treats, X) are restricted to drugs. This restriction 
eliminates trivial cross-type distinctions and makes 
prediction comparable across relation types, but it also 
narrows the negative set to a more challenging within-type 
candidate pool.

\section{Structural Priors: Implementation and Preprocessing Costs}
\label{app:structural-priors}

The first stage of the \sysname{} pipeline is a preprocessing step that constructs local and global structural priors,
both from the training triples only. The two priors are then frozen, and their fusion with the learned
embedding and any domain-specific features (Sec.~\ref{sec:fusion}) is trained
end-to-end with the downstream task.
Sec.~\ref{sec:exp-robustness} investigated the effect of the choice of prior on accuracy and justified our decision to use Bloom filters and TransE by default. Here, we provide more details on the implementation of the priors and examine the preprocessing cost required to obtain them.

\subsection{Bloom Filter: Construction and Parameter Selection}
\label{app:bloom}

\paragraph{Rationale.}
A KG node can learn about its 1-hop neighbors in a single round of
message passing under full-batch training, since the node receives
messages from \emph{all} its neighbors.  Full-batch training is
infeasible for large KGs, however, so mini-batch training randomly
samples neighbors, providing only a partial neighborhood view.

To mitigate this loss, we represent the set of 1-hop neighbors with a Bloom filter: Bloom
filters are fixed-length, tunable, accurate, cheap to compute, and
require no explicit adjacency storage.  Each node thus carries a
lightweight approximate summary of its full neighborhood,
independent of sampling during mini-batch formation.  The Bloom
filter provides a compact structural prior that the downstream GNN
must still interpret, and we show it is effective at recovering
the information that sampling discards.

\paragraph{Construction.}

We assign each node $i\in\mathcal{E}$ an $m$-bit Bloom filter
$B[i] \in \{0,1\}^m$ encoding its 1-hop neighborhood.  
The full set $\{B[i]\}_{i\in\mathcal{E}}$ is constructed in a single pass over the training triples $\mathcal{T}$ (Algorithm~\ref{alg:bloom}).

To accurately reflect node $i$'s neighborhood, $B[i]$ must encode
both outgoing edges (where $i$ is the head in $(i, r, t)$) and
incoming edges (where $i$ is the tail in $(h, r, i)$).  It must
also record the relation type, not only the neighbor's identity.
For outgoing edges we encode the pair as the string
\texttt{"r\_t"}; for incoming edges as \texttt{"r\_h"}.  For
every triple $(h, r, t)$ the algorithm therefore inserts
\texttt{"r\_t"} into $B[h]$ and \texttt{"r\_h"} into $B[t]$.
Inserting a string sets $k$ bit positions given by $k$ independent
hash functions applied to that string.  After processing all
triples, $B[i]$ holds a compact approximate encoding of node
$i$'s relational neighborhood without storing any adjacency list
explicitly (see Fig.~\ref{fig:bloom-transE} for an example).

\begin{wrapfigure}{r}{0.5\textwidth}   
\vspace{-1em}
\begin{algorithm}[H]
\small
\DontPrintSemicolon
\caption{\textsc{BuildBloomFilters}}
\label{alg:bloom}
\KwIn{$\mathcal{T}$ (triple list), $m$ (number of bits), $k$ hash functions}
\KwOut{$\{B[i]\}_{i \in \mathcal{E}}$ (Bloom filters for all nodes)}
\ForEach{$i \in \mathcal{E}$}{
    initialize $B[i] \leftarrow \mathbf{0}_{m}$\;
}
\ForEach{$(h, r, t) \in \mathcal{T}$}{
    $B[h].\textsc{insert}(\texttt{"r\_t"})$\;
    $B[t].\textsc{insert}(\texttt{"r\_h"})$\;
}
\Return{$\{B[i]\}$}\;
\end{algorithm}
\end{wrapfigure}

\vspace{-9pt}
\paragraph{Parameter selection.}

The key dataset-dependent parameter is $n$, the expected number of
elements inserted per filter.  A naive choice is the maximum
node degree, but degree distributions in KGs are highly skewed:
a few outlier nodes (e.g., papers with unusually large citation
counts in MAG240M) would force an unnecessarily large filter for
all others.  We instead take the sum of the 95th-percentile
out-degree and the 95th-percentile in-degree as a stable estimate
of the typical neighborhood size.
Given $n$ and a target false-positive rate $\varepsilon$, the
required bit width and hash count follow standard formulas~\cite{bloom-org}:

\[
  m = -\frac{n \ln \varepsilon}{(\ln 2)^2}, \qquad
  k = \frac{m}{n} \ln 2.
\]
A Bloom filter also becomes uninformative as its fill ratio
approaches 1.  The expected fill ratio for a node with $n$ edges
is
\[
  \rho(n;\,m,k) = 1 - e^{-kn/m}.
\]
We verify that $\rho(n_{\text{med}};\,m,k) \leq \rho^* = 0.5$,
where $n_{\text{med}}$ is the median node degree, to ensure
typical nodes remain discriminative.

In practice, we round $m$ up to the nearest 
convenient number
satisfying both criteria.  This yields $m = 128$ for OGB-MAG
(median degree $\approx 5$), $m = 500$ for WikiKG2 (median degree
$\approx 6$), and $m = 8{,}000$ for BioKG (median degree
$\approx 22$). 
Doubling or halving $m$ around these values
changes 
test accuracy by
$< 0.5$ points on OGB-MAG
and 
test MRR by $< 0.005$ on WikiKG2.
The training configuration is further discussed in Sec.~\ref{app:lp-hyperparams}.

When $m \leq 1024$, each filter is stored as a \texttt{uint32}
array.  For $m > 1024$ we apply run-length encoding and store all
filters in a single memory-mapped file, paging in only the rows
needed per mini-batch.  Storage footprints: OGB-MAG $\approx 30$\,MB;
WikiKG2 $\approx 156$\,MB; BioKG $\approx 94$\,MB---all comfortably
within CPU RAM.

\subsection{MinHash: Construction and Parameter Selection}
\label{app:minhash}

The \emph{MinHash signature} represents the same set of pairs through $k$ minimum-valued hashes, so two entities with similar neighborhoods produce signatures whose Jaccard similarity is preserved with low variance, which provides a different similarity 
summary than Bloom filters. 
For each entity $e$, we use $k$ independent hash functions and compute
\[
  f_{\mathrm{MinHash}}(e)_i
  = \min_{\sigma \in \widetilde{\mathcal{N}}(e)} h_i(\sigma).
\]
The empirical match rate over $k$ coordinates is an unbiased
estimator of the Jaccard similarity of two neighborhoods with
standard deviation $\sqrt{J(1-J)/k}$.  We use $k = 256$, giving
$\sigma \approx 0.031$ at $J = 0.5$.  In our ablations
(Tab.~\ref{tab:robustness-both}), Bloom and MinHash yield comparable downstream
accuracy.

\subsection{Relation Distribution (RelDist)}
\label{app:reldist}
The \emph{relation-distribution histogram} (RelDist) 
records for each relation $r \in \mathcal{R}$ the count of triples in which each entity $e$ participates with $r$, yielding a fixed-size $|\mathcal{R}|$-dimensional histogram.
For each entity $e$, form an $|\mathcal{R}|$-dimensional vector
where the $r$-th component counts $e$'s participations with
relation $r$ in both head and tail roles.  Construction time is
$\mathcal{O}(|\mathcal{T}|)$.  The vector is sparse and is
projected to $d$ dimensions via 
an MLP
before feature fusion.
RelDist preserves exact relational role counts but discards
neighbor identity, limiting its ability to distinguish entities
that play the same relational roles with different partners.
The global feature partly compensates for this.

\paragraph{Complexity summary of local priors:}
\begin{itemize}[noitemsep]
  \item Bloom filter: $m$ bits/entity, $\mathcal{O}(k|\mathcal{T}|)$ time.
  \item MinHash: $k$ integers/entity, $\mathcal{O}(k|\mathcal{T}|)$ time.
  \item RelDist: $|\mathcal{R}|$ counts/entity (sparse),
        $\mathcal{O}(|\mathcal{T}|)$ time.
\end{itemize}

\subsection{Global Prior: KGE Pre-Training Configuration}
\label{app:global-prior}

The global prior consists of frozen entity vectors from a KGE model
trained on the full training set $\mathcal{T}$ before any
downstream training begins.  TransE is the default since its
translational geometry produces entity coordinates whose distances
directly reflect relational proximity, a natural input for the
fusion gate.
ComplEx is used for BioKG to handle symmetric
relations.
The pre-training configuration for the KGEs is given in Tab.~\ref{tab:app-kge-config}.

\begin{table}[ht]
\centering
\caption{KGE pre-training configuration for different datasets.}
\label{tab:app-kge-config}
\begin{tabular}{l|cccc}
\toprule
Dataset  & Dim & Epochs  & Model \\
\midrule
OGB-MAG  & 128 & 30  & TransE  \\
MAG240M  & 128 & 30  & TransE  \\
WikiKG2  & 250 & 30  & TransE  \\
BioKG    & 1000 & 30  & ComplEx \\
\bottomrule
\end{tabular}
\end{table}

\subsection{Preprocessing Costs}
\label{app:preprocessing-costs}

Both preprocessing steps---computing the local and the global priors---are one-time costs amortized across all
downstream training runs, seeds, and backbones.
The same Bloom
filters and KGE embeddings are reused in all experiments.
Tab.~\ref{tab:app-preprocessing} shows the preprocessing time and memory requirements for the OGB datasets.
Preprocessing time is proportional to dataset size, and most of it is spent on KGE embeddings ($\sim$2\,h for WikiKG2 and $\sim$6.5\,h for MAG240M, on one GPU).
This is a modest one-time overhead given that
GNN training itself runs for many more hours.

\begin{table}[ht]
\centering
\caption{gHAWK preprocessing costs on OGB datasets.
}
\label{tab:app-preprocessing}
\begin{tabular}{ll|rr}
\toprule
Dataset & Component & Wall-clock time & Peak RAM \\
\midrule
\multirow{2}{*}{OGB-MAG (1.94M / 21M)}
  & Bloom (128\,b)  & 14.8\,s        & 6.1\,GiB \\
  & TransE (128-d)  & 24\,min 46\,s  & 5.1\,GiB \\
\midrule
\multirow{2}{*}{MAG240M (244M / 1.3B)}
  & Bloom (128\,b)  & $\sim$3\,min   & $\sim$24\,GiB \\
  & TransE (128-d)  & $\sim$6.5\,h   & $\sim$48\,GiB \\
\midrule
\multirow{2}{*}{WikiKG2 (2.5M / 16.1M)}
  & Bloom (500\,b)  & 38\,s          & 7.2\,GiB \\
  & TransE (250-d)  & $\sim$2\,h     & 8.4\,GiB \\
\midrule
\multirow{2}{*}{BioKG (94K / 4.8M)}
  & Bloom (8000\,b) & 12\,s          & 0.6\,GiB \\
  & ComplEx (1000-d) & $\sim$18\,min  & 1.2\,GiB \\
\bottomrule
\end{tabular}
\end{table}

For DBLP, ACM, and IMDb, total preprocessing is under five minutes
per dataset on one GPU, with TransE pre-training dominating
the one-time cost. Tab.~\ref{tab:app-preprocessing-hgb} shows the details.

\begin{table}[ht]
\centering
\caption{gHAWK preprocessing costs on HGB datasets.
}
\label{tab:app-preprocessing-hgb}
\begin{tabular}{ll|rr}
\toprule
Dataset & Component & Wall-clock time & Peak RAM \\
\midrule
\multirow{2}{*}{DBLP (26K / 239K)}
  & Bloom (256\,b)   & 1.4\,s        & 1.38\,GiB \\
  & TransE (512-d)   & 2\,min\,19\,s  & 1.51\,GiB \\
\midrule
\multirow{2}{*}{ACM (10K / 547K)}
  & Bloom (256\,b)   & 3.2\,s        & 1.17\,GiB \\
  & TransE (512-d)   & 4\,min\,31\,s  & 1.30\,GiB \\
\midrule
\multirow{2}{*}{IMDb (21K / 86K)}
  & Bloom (256\,b)   & 0.5\,s        & 1.21\,GiB \\
  & TransE (512-d)   & 51.1\,s        & 1.31\,GiB \\
\bottomrule
\end{tabular}
\vspace{-2pt}
\end{table}

\section{Training and Inference Cost}
\label{app:efficiency}

Sec.~\ref{sec:exp-efficiency} summarizes \sysname{}'s cost; here we give the per-backbone measurements. For each backbone, the \sysname{} priors are toggled with the decoder, optimizer, batch size, and negative sampling held fixed, so the reported differences reflect the priors alone.
Tab.~\ref{tab:efficiency-node} reports node property prediction cost on OGB-MAG
and
Tab.~\ref{tab:efficiency-lp} reports link prediction cost on WikiKG2.

\begin{table}[ht]
\centering
\small
\caption{Node property prediction cost on OGB-MAG, toggling the \sysname{} priors with default Word2Vec features held fixed.
$^\ast$R-GCN is trained full-batch, which accounts for its short training time; the other backbones use mini-batch sampling.}
\label{tab:efficiency-node}
\setlength{\tabcolsep}{8pt}
\begin{tabular}{l|cc|c}
\toprule
& \multicolumn{2}{c|}{Total train time} & Peak GPU \\
Backbone & Default & \sysname{} & (GiB) \\
\midrule
R-GCN$^\ast$ & 9.2\,min  & 9.7\,min  & 27.1 \\
GraphSAGE  & 7.94\,h   & 9.48\,h   & 23.7 \\
GraphSAINT & 1.52\,h   & 1.38\,h   & 11.3 \\
ClusterGCN & 1.03\,h   & 1.18\,h   & 22.6 \\
GAT        & 23.89\,h  & 25.16\,h  & 38.8 \\
HGT        & 21.82\,h  & 21.62\,h  & 8.2  \\
\bottomrule
\end{tabular}
\end{table}

\begin{table}[ht]
\centering
\small
\caption{Link prediction cost on WikiKG2, toggling the \sysname{} priors on
each matched backbone. Per-epoch time is mean wall-clock training time over
all epochs under the protocol of Tab.~\ref{tab:lp-hyperparams}; overhead is
the \sysname{}/Default ratio. Test MRR is repeated from
Tab.~\ref{tab:wikikg2-main}. $^\ast$R-GCN uses batch size 512
(Tab.~\ref{tab:lp-hyperparams}), so each epoch has twice as many steps.}
\label{tab:efficiency-lp}
\setlength{\tabcolsep}{5pt}
\begin{tabular}{l|ccc|cc}
\toprule
& \multicolumn{3}{c|}{Time (min/epoch)} & \multicolumn{2}{c}{Test MRR} \\
Backbone & Default & \sysname{} & Overhead & Default & \sysname{} \\
\midrule
RotatE (decoder-only) & 43.6  & 53.4  & $1.22\times$ & 0.4341 & \textbf{0.6941} \\
GraphSAGE             & 41.5  & 60.7  & $1.46\times$ & 0.5689 & \textbf{0.6129} \\
R-GCN$^\ast$          & 326.6 & 343.2 & $1.05\times$ & 0.3851 & \textbf{0.5957} \\
CompGCN               & 33.5  & 35.9  & $1.07\times$ & 0.6256 & \textbf{0.6562} \\
\bottomrule
\end{tabular}
\end{table}

On OGB-MAG, total training time barely moves and is lower with \sysname{} on
some backbones, since the priors are read from CPU and disk rather than
recomputed. On WikiKG2, per-epoch time rises by 5--46\% on GNN backbones and
by 22\% in the decoder-only pipeline (Tab.~\ref{tab:efficiency-lp}).

Peak GPU memory is determined by the backbone rather than the priors, because
the priors are moved to the GPU only for the entities being processed: the
sampled mini-batch for most backbones, and the whole graph for full-batch
R-GCN on OGB-MAG, where the priors are still small next to the backbone's own
memory footprint. The node property prediction values are shown in Tab.~\ref{tab:efficiency-node}. For link prediction on WikiKG2, peak GPU memory is 29.7\,GiB with \sysname{} versus 28.4\,GiB without for
RotatE ($+4.6\%$).

On a GNN backbone the fusion module adds $0.03\%$--$0.07\%$ of parameters.
In
the decoder-only pipeline it reaches $2.5\%$.
The structural features are stored on disk and paged per
mini-batch ($2.66$\,GB on WikiKG2), so they never occupy GPU memory, and
inference tracks the backbone's own forward pass because the fused
representation is frozen after training and materialized once.

\section{Node Property Prediction: Extended Results}
\label{app:nodeprop}

\subsection{Training Configuration}
\label{app:nodeprop-config}

 The GNNs are trained according to the following configurations:

\textbf{OGB-MAG / MAG240M.}  Two GNN layers; hidden dimension
256; dropout 0.5; fanout $[25, 20]$; AdamW with learning rate
$10^{-3}$; early stopping on validation accuracy. R-GCN on OGB-MAG is trained full-batch; all other runs use mini-batch sampling.

\textbf{DBLP / ACM / IMDb.}  GraphSAGE, two layers, hidden
dimension 256, dropout 0.5, fanout $[20, 20]$, AdamW with cosine
schedule, batch size 512, learning rate $10^{-3}$, weight decay
$10^{-4}$, 80 epochs evaluated every 5 epochs, 5 seeds.  Bloom
uses 256 bits, $k = 5$ hash functions; MinHash uses 128
permutations; KGE methods use dimension 512 and 30 epochs;
Node2Vec uses dimension 64, walk length 40, context size 10, 10
walks per node, 5 negative samples.

\subsection{Extended Backbones on OGB-MAG}
\label{app:nodeprop-extra-backbones}

Beyond the six backbones in the main paper, we evaluated four
additional architectures on OGB-MAG: a decoder-only MLP (no
message passing at all), SGC~\cite{sgc} (a linear graph
convolution), SIGN~\cite{sign} (precomputed spectral
propagation), and SGFormer~\cite{sgformer} (a global-attention
transformer). Tab.~\ref{tab:app-extended-backbones} reports test accuracy
with the default Word2Vec features and with \sysname{} added on
top.

\begin{table}[ht]
\centering
\caption{Extended GNN backbones on OGB-MAG (test accuracy \%,
default Word2Vec features).}
\label{tab:app-extended-backbones}
\small
\begin{tabular}{l|ccc}
\toprule
Model    & Default & \sysname{} & $\Delta$ \\
\midrule
MLP      & 31.95 & \textbf{35.61} & $+3.66$ \\
SGC      & 38.42 & \textbf{40.84} & $+2.42$ \\
SIGN     & 37.54 & \textbf{40.46} & $+2.92$ \\
SGFormer & 51.35 & \textbf{56.46} & $+5.11$ \\
\bottomrule
\end{tabular}
\end{table}

The most striking row is the MLP. A decoder-only MLP equipped
with \sysname{} features---no message passing whatsoever---reaches
$35.61\%$ on OGB-MAG, exceeding the Text-only GraphSAGE baseline
in Tab.~\ref{tab:node-default} ($31.66\%$) by $+4.0$ points.
SGC, SIGN, and SGFormer all benefit consistently from \sysname{},
with SGFormer gaining the most ($+5.11$ points) because its
global-attention mechanism lacks an explicit local inductive
bias and therefore benefits more from the Bloom filter's
neighborhood signature.

\subsection{Ablation on Default Text Embedding Features}
\vspace{-4pt}
\label{app:nodeprop-no-text}

The default features used in Tab.~\ref{tab:node-default} are
informative but expensive. RoBERTa, in particular, produces
$768$-dimensional vectors that dominate first-layer GNN memory
and require running an external language model over every paper
in the graph. A natural follow-up question is whether
\sysname{}'s structural priors can substitute for these
features, not just complement them.
Tab.~\ref{tab:node-notext} reports the same six backbones 
without using the default text embedding features of the two dataset.
\emph{No Text} is initialized using random vectors so the
model relies only on graph structure, \emph{Local}
adds the Bloom filter, and \emph{\sysname{}} adds both
Bloom and TransE.

\begin{table*}[ht]
\centering
\caption{Node property prediction accuracy (\%) without using the domain-specific 
features. \emph{No Text} uses random vectors as input;
\emph{Local} adds the Bloom filter; \emph{\sysname{}} means both
Bloom and TransE features.}
\label{tab:node-notext}
\small
\setlength{\tabcolsep}{6pt}
\begin{tabular}{l|ccc|ccc}
\toprule
& \multicolumn{3}{c|}{OGB-MAG} & \multicolumn{3}{c}{MAG240M} \\
Model & No Text  & Local & \sysname{}
      & No Text & Local & \sysname{} \\
\midrule
R-GCN      & 23.37 & 42.88 & \textbf{43.71}
           & 21.66 & 49.32 & \textbf{59.23} \\
GraphSAGE  & 16.14 & 37.42 & \textbf{40.69}
           & 23.82 & 52.88 & \textbf{61.52} \\
GraphSAINT & 27.41 & 48.57 & \textbf{54.55}
           & 27.34 & 60.90 & \textbf{60.98} \\
ClusterGCN & 22.12 & 39.01 & \textbf{43.89}
           & 25.19 & 54.71 & \textbf{59.82} \\
GAT        & 24.12 & 32.83 & \textbf{45.34}
           & 26.15 & 54.31 & \textbf{62.15} \\
HGT        & 25.65 & 44.14 & \textbf{48.94}
           & 28.45 & 49.81 & \textbf{57.55} \\
\bottomrule
\end{tabular}
\end{table*}

Without any features, all six backbones produce less than $30\%$
accuracy on both datasets, confirming that mini-batch training
on graph structure alone loses too much information for accurate
prediction even when the backbone explicitly models relation
types. Adding the Bloom filter alone lifts every backbone by
$9$--$21$ percentage points on OGB-MAG and by $21$--$34$ points on
MAG240M, recovering most of the gap with the default-features
baselines. Adding TransE on top further improves accuracy on
every backbone. This demonstrates the importance of both local and global priors in \sysname{}.

The most striking observation is what
happens when we compare across Tab.~\ref{tab:node-default} and 
Tab.~\ref{tab:node-notext}.
On OGB-MAG, the
structural-only \emph{\sysname{}} configuration in
Tab.~\ref{tab:node-notext} beats the \emph{Default}
(Word2Vec-only) column in Tab.~\ref{tab:node-default} for every
single backbone, e.g., GraphSAINT reaches $54.55\%$ with no
default features versus $46.81\%$ with Word2Vec alone. On
MAG240M, where the default features are RoBERTa, which are more expressive than Word2Vec albeit more expensive, structural-only
\sysname{} recovers roughly $90\%$ of RoBERTa-only accuracy on
most backbones while using a fraction of the dimensionality.

Comparing across Tab.~\ref{tab:node-default} and 
Tab.~\ref{tab:app-extended-backbones}, we see that a decoder-only MLP equipped with \sysname{} features and no
message passing reaches $35.61\%$ on OGB-MAG
(Tab.~\ref{tab:app-extended-backbones}), exceeding
the $31.66\%$ of GraphSAGE with the default features only (Tab.~\ref{tab:node-default}).
A model
with no message passing and no text embedding features but with structural priors outperforms a full
message-passing GNN that has text embedding features but no structural priors.

As we argued in the
introduction, when every node carries an explicit representation
of graph structure, training can compensate for the information
loss that mini-batch sampling causes on large KGs, and \sysname{}
is not adding noise on top of strong features but providing real
structural information that the GNN would otherwise need to
reconstruct from scratch.

\vspace{-4pt}
\subsection{Convergence Time of GNN Backbones}
\label{app:nodeprop-convergence}
\begin{figure}[ht]
    \centering
    \includegraphics[width=0.90\linewidth]{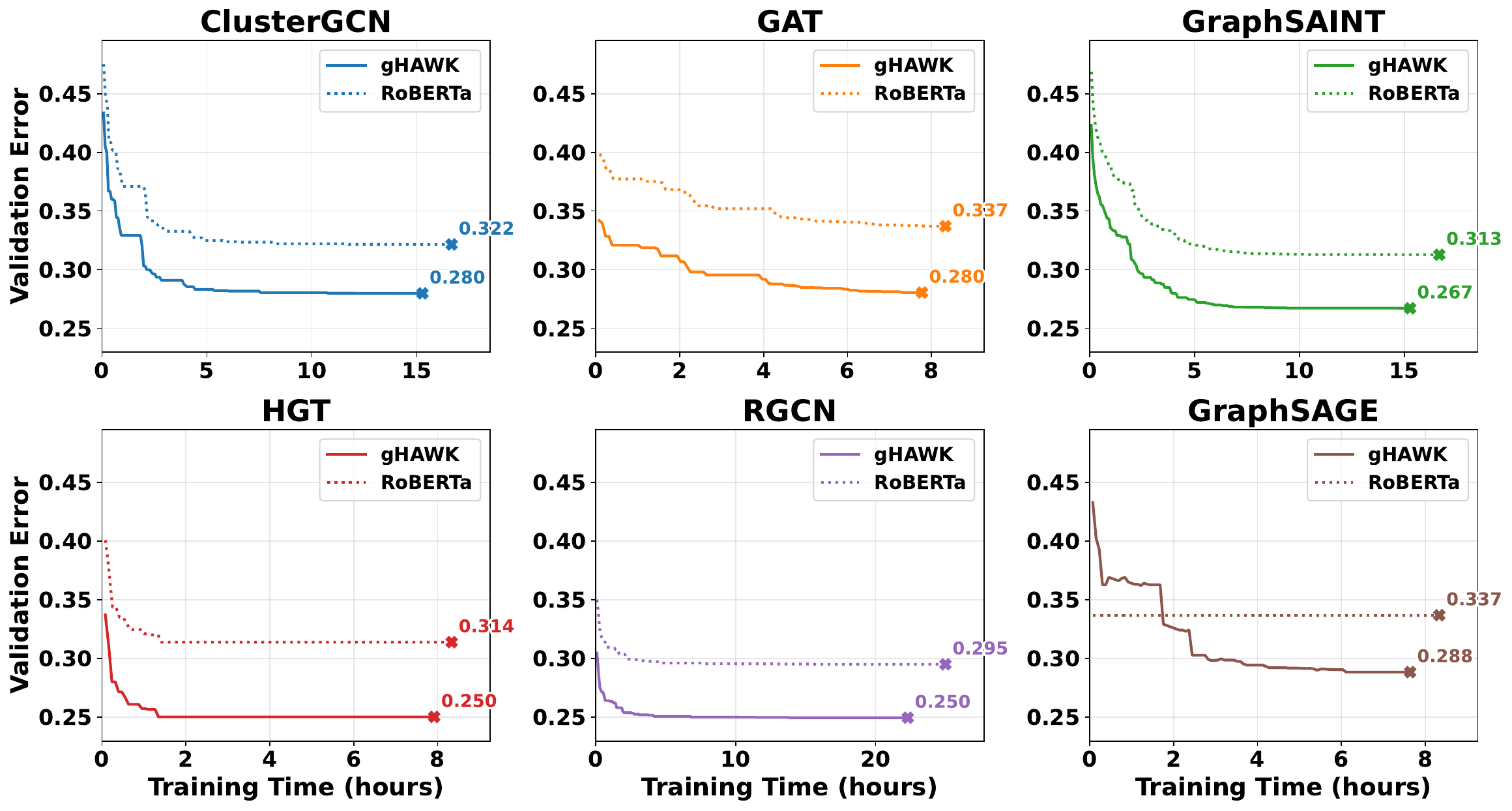}
    \caption{Model convergence on MAG240M (single run per configuration). The plots show validation error versus training time (hours) for each GNN backbone. Dotted lines represent RoBERTa features only, and solid lines represent RoBERTa+\sysname{}. 
    X markers denote the final validation error, with the numeric label showing its value.}
    \label{fig:mag240m-convergence}
\end{figure}

\sysname{} also affects how well the GNN converges.
Fig.~\ref{fig:mag240m-convergence} plots validation error versus training
time on MAG240M for every backbone, comparing the default RoBERTa features
(dotted) to RoBERTa + \sysname{} (solid). \sysname{} reaches a lower final
validation error than the RoBERTa-only baseline on every backbone, and stays
below it for most of training. Part of the early-training advantage comes
from the fact that \sysname{} projects the $768$-dimensional RoBERTa vector
to $384$ dimensions before fusion, which reduces the model's parameter count
by $2$--$12\%$ across backbones (e.g., GAT $3.1 \to 2.8$M;
GraphSAGE/GraphSAINT/ClusterGCN $4.9 \to 4.4$M; HGT $31.4 \to 30.7$M; R-GCN
$34.2 \to 30.1$M), so each gradient step is cheaper. The persistent gap in
final validation error reflects structural information that the RoBERTa
baseline does not recover, even after extended training.

\vspace{-9pt}
\subsection{Sensitivity to \sysname{} Feature Size}
\label{app:nodeprop-hyperparam}

Tab.~\ref{tab:app-hyperparam} reports test accuracy on OGB-MAG
as a function of the number of bits in the Bloom filter and the number of dimensions in TransE, using
GraphSAGE with the default Word2Vec features.

\begin{table}[ht]
\centering
\caption{OGB-MAG test accuracy (\%, GraphSAGE + Word2Vec) for different sizes of the Bloom filter and TransE embedding. Single run result.}
\label{tab:app-hyperparam}
\begin{tabular}{l|ccc}
\toprule
TransE dim $\backslash$ Bloom bits & 128 & 256 & 512 \\
\midrule
128 & 43.04 & 41.47 & 49.08 \\
256 & 44.15 & 42.41 & 41.84 \\
512 & 44.68 & 43.58 & 42.96 \\
\bottomrule
\end{tabular}
\end{table}

Accuracy generally increases with TransE dimension for 128- and 256-bit filters. The Bloom
filter shows a non-monotonic pattern under GraphSAGE: 128 bits
outperforms 256 and 512, except that 512 bits paired with 128-d
TransE gives the best accuracy in this single run.  Under mini-batch sampling, the smaller
filter may introduce a stochastic regularization effect that benefits
GraphSAGE, while larger deterministic filters cannot be fully
exploited because each mini-batch accesses only a partial
neighborhood.  The default 128-bit / 128-d configuration is the most
efficient.

\subsection{Feature Fusion Strategies}
\label{app:nodeprop-fusion}

Tab.~\ref{tab:app-fusion} compares strategies for feature fusion:
simple concatenation (\textbf{C}), concatenation with
element-wise (Hadamard) products (\textbf{CH}), and the learned scalar
gating with residual connection (\textbf{GR}) used by default.

\begin{table}[ht]
\centering
\caption{Fusion strategy comparison on OGB-MAG (test accuracy\,\%).
C: Concatenation; CH: Concat-Hadamard; GR: Gated Residual. Single run result.}
\label{tab:app-fusion}
\begin{tabular}{l|cccl}
\toprule
Backbone   & C     & CH    & GR             & Best \\
\midrule
ClusterGCN & 43.18 & 44.89 & \textbf{45.70} & GR \\
GraphSAINT & 47.31 & 48.49 & \textbf{57.97} & GR \\
GraphSAGE  & 31.43 & 37.16 & \textbf{43.04} & GR \\
MLP        & \textbf{35.61} & 29.93 & 29.39 & C  \\
SGC        & 38.99 & 37.84 & \textbf{40.84} & GR \\
SGFormer   & \textbf{56.46} & 53.71 & 54.41 & C  \\
\bottomrule
\end{tabular}
\end{table}

Gated Residual is the consistent winner for message-passing
backbones.  The gain over concatenation is largest on GraphSAINT
(47.31 $\to$ 57.97), where the gate learns to suppress structural
channels that are noisy for a given subgraph partition.  For
non-message-passing (MLP) and global-attention (SGFormer)
backbones, simple concatenation wins; these architectures already
learn implicit attention over their inputs and an additional gate
adds redundant capacity.

\section{Link Prediction: Extended Results}
\label{app:linkpred}

\subsection{Training Configuration}
\vspace{-2pt}
\label{app:lp-hyperparams}
Tab.~\ref{tab:lp-hyperparams} reports the training configuration for the
link prediction experiments on WikiKG2. 
\begin{table}[h]
\centering
\small
\caption{Training configuration for WikiKG2 link prediction. All backbones
use these values except where R-GCN is noted.}
\label{tab:lp-hyperparams}
\begin{tabular}{ll}
\toprule
Hyperparameter & Value \\
\midrule
\multicolumn{2}{l}{\emph{Optimization}} \\
\quad Batch size              & 1024 (R-GCN: 512) \\
\quad Optimizer               & AdamW \\
\quad Learning rate           & $1\mathrm{e}{-3}$ \\
\quad Weight decay            & $1\mathrm{e}{-5}$ \\
\quad Dropout                 & 0.1 \\
\quad Training steps          & up to 100 epochs, early stopped \\
\quad Early stopping          & validation MRR \\
\addlinespace
\multicolumn{2}{l}{\emph{Loss and negative sampling}} \\
\quad Loss                    & self-adversarial~\citep{rotate} \\
\quad Self-adv.\ temperature $\alpha$ & 1.0 \\
\quad Margin $\gamma$         & 12 \\
\quad Negatives per positive $K$ & 32 (R-GCN: 6) \\
\quad Negative sampler        & uniform \\
\addlinespace
\multicolumn{2}{l}{\emph{Embeddings and structural features}} \\
\quad Entity / KGE dimension  & 250 \\
\quad Bloom filter bits $m$   & 500 \\
\quad Bloom hash functions $k$ & 3 \\
\quad Global feature          & TransE \\
\addlinespace
\multicolumn{2}{l}{\emph{Fusion module}} \\
\quad Projection MLPs         & 2 layers, 256-dim hidden \\
\quad Fusion MLP              & 2 layers, 512-dim output \\
\addlinespace
\multicolumn{2}{l}{\emph{GNN backbones}} \\
\quad GNN layers              & 2 \\
\quad GNN hidden dim          & 256 (R-GCN: 100) \\
\quad Sampling fanout         & [25, 20] (R-GCN: [10, 5])\\
\bottomrule
\end{tabular}
\end{table}

All four configurations (RotatE and
three GNN backbones) share these hyperparameters, with one exception. R-GCN,
whose parameter count grows with the 535 relation types (Sec.~\ref{sec:backbones}),
uses a batch size of 512 and a hidden dimension of 100 so that training fits
in the memory of a single 48\,GB NVIDIA L40S GPU. Both R-GCN runs, with and
without \sysname{}, use these settings. Within each backbone, the runs with
and without \sysname{} use identical settings, so the differences in
Tab.~\ref{tab:wikikg2-main} reflect the node features rather than tuning.

BioKG's configuration inherits most of these values, with two exceptions:
the global feature is ComplEx, which encodes the symmetric relations that
dominate BioKG (Sec.~\ref{sec:exp-link-biokg}), and the Bloom filter uses more bits.
BioKG's median entity degree is $22$, with a heavy upper tail: a non-trivial
fraction of evaluation queries involve entities with degree $>100$. Holding
the saturation budget of the Bloom filter $\rho^* = 0.5$ at the median is insufficient for these
high-degree entities. We size the filter at $8{,}000$ bits, which keeps
saturation below budget at the 90th percentile of the degree distribution.
Total storage for the Bloom filters is approximately $94$\,MB, and downstream
test MRR is insensitive to perturbations within $\pm 50\%$ of this size.

\subsection{Detailed \sysname{} Link Prediction Results}
\label{app:lp-wikikg2-detailed}

Tab.~\ref{tab:wikikg2-detailed} and~\ref{tab:biokg-detailed}
report filtered MRR, Hits@3, and Hits@10 on validation and test
sets for all \sysname{} configurations.
The advantage provided by \sysname{} is evident in both tables and across all metrics.

\begin{table}[ht]
\centering
\footnotesize
\setlength{\tabcolsep}{3pt}
\caption{Detailed link prediction results on WikiKG2: mean $\pm$ standard
deviation over five seeds. Bold: best per column.}
\label{tab:wikikg2-detailed}
\begin{tabular}{@{}lcccccc@{}}
\toprule
 & \multicolumn{3}{c}{Validation} & \multicolumn{3}{c}{Test} \\
\cmidrule(lr){2-4} \cmidrule(l){5-7}
Method & MRR & H@3 & H@10 & MRR & H@3 & H@10 \\
\midrule
RotatE
  & 0.4352\std{.0004} & 0.4472\std{.0006} & 0.5009\std{.0005}
  & 0.4341\std{.0009} & 0.4411\std{.0006} & 0.4907\std{.0011} \\
\quad + \sysname{}
  & \textbf{0.7032}\std{.0014} & \textbf{0.7425}\std{.0011} & \textbf{0.8441}\std{.0017}
  & \textbf{0.6941}\std{.0016} & \textbf{0.7328}\std{.0019} & \textbf{0.8443}\std{.0023} \\
\addlinespace
GraphSAGE
  & 0.5720\std{.0060} & 0.6169\std{.0066} & 0.7528\std{.0076}
  & 0.5689\std{.0040} & 0.6025\std{.0041} & 0.7391\std{.0035} \\
\quad + \sysname{}
  & 0.6216\std{.0061} & 0.6654\std{.0042} & 0.8060\std{.0028}
  & 0.6129\std{.0024} & 0.6453\std{.0028} & 0.7884\std{.0021} \\
\addlinespace
R-GCN
  & 0.3717\std{.0233} & 0.4136\std{.0282} & 0.5702\std{.0255}
  & 0.3851\std{.0225} & 0.4309\std{.0250} & 0.6091\std{.0240} \\
\quad + \sysname{}
  & 0.5953\std{.0031} & 0.6373\std{.0049} & 0.7879\std{.0069}
  & 0.5957\std{.0039} & 0.6310\std{.0036} & 0.7853\std{.0046} \\
\addlinespace
CompGCN
  & 0.6275\std{.0033} & 0.6740\std{.0036} & 0.8199\std{.0022}
  & 0.6256\std{.0039} & 0.6665\std{.0035} & 0.8142\std{.0035} \\
\quad + \sysname{}
  & 0.6661\std{.0016} & 0.7060\std{.0014} & 0.8352\std{.0014}
  & 0.6562\std{.0010} & 0.6950\std{.0012} & 0.8336\std{.0018} \\
\bottomrule
\end{tabular}
\end{table}

\begin{table}[ht]
\centering
\small
\caption{Detailed link prediction results on BioKG.}

\label{tab:biokg-detailed}
\setlength{\tabcolsep}{4pt}
\begin{tabular}{l|ccc|ccc}
\toprule
& \multicolumn{3}{c|}{Validation} & \multicolumn{3}{c}{Test} \\
Method & MRR & H@3 & H@10 & MRR & H@3 & H@10 \\
\midrule
RotatE                       & 0.7997 & 0.8577 & 0.9399 & 0.7989 & 0.8588 & 0.9405 \\
\textbf{RotatE + \sysname{}} & \textbf{0.8380} & \textbf{0.8860} & \textbf{0.9496} & \textbf{0.8378} & \textbf{0.8849} & \textbf{0.9487} \\
\bottomrule
\end{tabular}
\end{table}

\subsection{Single-Prior Ablation}
\label{app:lp-single-prior}

The robustness grid in Sec.~\ref{sec:exp-robustness} shows that
all combinations of local and global priors converge to
similar 
MRR values. Tab.~\ref{tab:lp-single-prior} extends this
analysis with single-prior baselines on WikiKG2 (RotatE decoder-only
configuration), showing that each prior contributes
substantially on its own and the two carry complementary
information.

\begin{table}[ht]
\centering
\small
\caption{Single-prior ablation on WikiKG2 (RotatE decoder-only).
Each structural prior alone contributes substantially over the
RotatE baseline, and the combination yields the full gain. Single run result}
\label{tab:lp-single-prior}
\begin{tabular}{ll|cc}
\toprule
Local & Global & Test MRR & Head MRR \\
\midrule
--    & --     & 0.4345 & 0.0868 \\
Bloom & --     & 0.6731 & 0.3713 \\
--    & TransE & 0.6641 & 0.3791 \\
\midrule
Bloom & TransE & \textbf{0.6930} & \textbf{0.4627} \\
\bottomrule
\end{tabular}
\end{table}

Bloom alone takes test MRR from $0.4345$ to $0.6731$ ($+23.9$
points) and head MRR from $0.087$ to $0.371$ ($+28.4$ points).
TransE alone takes test MRR to $0.6641$ ($+23.0$ points) and
head MRR to $0.379$ ($+29.2$ points). The combination
(Bloom + TransE) reaches test MRR $0.6930$ and head MRR
$0.4627$. Neither prior subsumes the other: the local feature
captures neighborhood composition, the global feature captures
relational position, and the two carry complementary
information that adds beyond either alone.
\subsection{Using TransE as a Negative Sampler}
\label{app:neg-sampler}

\sysname{} samples negatives uniformly and relies on the self-adversarial
loss~\citep{rotate} to reweight them by difficulty. Because \sysname{}'s
preprocessing already trains a KGE for the global prior, the same KGE could
also mine \emph{hard} negatives, a role that usually requires training a
separate scoring model. We test whether this second role adds accuracy.

Tab.~\ref{tab:kge-roles} shows the effect of using the KGE as a global prior
and as a negative sampler, in the decoder-only RotatE configuration. The first
column uses \sysname{}'s default uniform negative sampler. The second column
uses TransE to score the candidate negatives and trains the model on the
negatives that TransE scores as hardest. In each column, the structural prior
is either Bloom + TransE or Bloom alone. Thus, the table shows the four
possible settings of using (or not using) TransE as a prior and as a negative
sampler.

Comparing the two columns of the table, we see that switching the sampler
changes test MRR by only $+0.002$ with Bloom + TransE, lowers it with Bloom
alone, and lowers validation MRR in both settings, likely because the
self-adversarial loss already focuses training on hard negatives. In contrast, comparing the rows of the table, we see that removing
TransE as a global prior costs $0.020$ to $0.028$ under either sampler. Thus,
using the KGE as a global prior provides a clear gain, while using it as a
negative sampler does not. These results justify \sysname{}'s design decision
to use the KGE only as a global prior.

\begin{table}[h]
\centering
\small
\caption{Using the KGE as a global prior and as a negative sampler on WikiKG2
(decoder-only RotatE, single run; validation / test MRR). Test MRR in the
uniform column matches Tab.~\ref{tab:lp-single-prior}.}
\label{tab:kge-roles}
\begin{tabular}{l|cc}
\toprule
& Uniform negatives & TransE-mined negatives \\
\midrule
Bloom + TransE & 0.7026 / 0.6930 & 0.7017 / 0.6954 \\
Bloom only     & 0.6851 / 0.6731 & 0.6762 / 0.6676 \\
\bottomrule
\end{tabular}
\end{table}

\subsection{Per-Entity vs.\ Per-Query Distributions}
\label{app:lp-stratification}

Tab.~\ref{tab:wikikg2-main} shows that \sysname{} produces an MRR gain of $+26$ percentage points on WikiKG2.
This large gain reflects a distinction between two ways of looking at the long tail. 
While $80\%$ of WikiKG2's entities appear in fewer than $10$ training triples, the distribution of test \emph{queries} is more concentrated: $92\%$ of test queries involve entities with between $1$ and $20$ training triples. 
These are the queries that benefit the most from the Bloom filter's ability to capture the long tail of the data distribution.
Since reporting MRR uniformly weights queries rather than entities, the gain reflects the query distribution rather than the entity distribution. 
The majority of the queries are in a regime where gradient-based embeddings cannot produce useful signal, and these queries benefit greatly from \sysname{}'s Bloom filter, so the overall gain in MRR is high.

A subtle exception is worth flagging. In the bucket of highest-frequency
entities ($1000+$ training triples, $0.21\%$ of queries), \sysname{}
underperforms RotatE ($0.7380$ vs.\ $0.8989$ head MRR). These entities
receive enough gradient signal to learn strong embeddings on their own, so
fusing in lower-capacity structural features dilutes them. Because the
bucket is so small, the effect on aggregate head MRR is negligible (about
$0.0003$), and it does not arise with R-GCN, which improves in every
frequency bucket, including $1000+$ (Tab.~\ref{tab:head-mrr-frequency-rgcn}).
This is consistent with our framework's premise: structural priors help most
where learned embeddings cannot reach, which is the long tail that dominates
the query distribution. 

\subsection{Doppelganger Confusion and Scoring with a Doppelganger Oracle}
\label{app:lp-oracle}

\paragraph{Empirical scope of doppelganger confusion.}
In Sec.~\ref{sec:exp-robustness}, we observed that the main source of error
in head prediction, i.e., $(?, r, t)$ queries, is confusion between the
correct head and candidates whose training neighborhoods overlap with it.
Consider a test query $(?, r, t)$ whose correct answer is $h_1$, and a
candidate $h_2$ for which $(h_2, r, t)$ is not a known triple, so that $h_2$
remains a negative under the filtered protocol.
If $h_1$ and $h_2$ share
training pairs, e.g., both $(h_1, r', t')$ and $(h_2, r', t')$ are in
$\mathcal{T}$, their neighborhoods look alike, and a model that ranks $h_2$
above $h_1$ lowers the MRR.
We call such candidates \emph{doppelgangers} of
the correct head. In the empirical analysis below, a candidate counts as a
doppelganger of the correct head $h^*$ if it shares at least one training
$(r, t)$ pair with $h^*$.

Detailed error analysis on WikiKG2 reveals that doppelganger confusion is the
dominant failure mode in head prediction. Of \sysname{}'s wrong predictions
at rank~$1$, $93\%$ share at least one training $(r, t)$ pair with the
correct head. The problem is not concentrated in a few pathological queries:
averaging across all test queries with masked heads, $240$ of the $500$
candidate negatives ($48\%$) are doppelgangers of the correct head. Because
the head distribution is the long tail of WikiKG2
(Sec.~\ref{sec:exp-link-headline}), the asymmetry is severe: head MRR for
RotatE 
is $0.087$ while tail MRR is $0.78$.
Doppelganger confusion is a head-side phenomenon; the $+0.47$ headroom above
\sysname{}'s head MRR ($0.463$) that we characterize below is exclusive to
head prediction.

\paragraph{Mechanism of doppelganger confusion.}
The phenomenon is a structural consequence of how query-independent
representations are built. Two entities $e_1$ and $e_2$ that share a training
pair $(r', t')$ receive the same contribution from it: the same bits in their
Bloom filters, the same count in RelDist, and, during gradient training, the
same pressure to score $(\cdot, r', t')$ well. The fewer training pairs they
do \emph{not} share, the closer their representations become, and at query
time the scoring function sees two similar representations that it cannot
reliably tell apart. Proposition~A (App.~\ref{app:proofs}) makes this
precise for every class of query-independent feature we consider, and
Corollary~A carries it over to the decoder's scores.

\paragraph{Scoring with a doppelganger oracle.}
Assume that we have an oracle that tells us the correct head $h^*$ for each
test query $(?, r, t)$. We define an \emph{oracle score} for each candidate
$c \neq h^*$ as
\[
  \hat{s}_{\mathrm{oracle}}(c) = \hat{s}(c) - \alpha \cdot S(c, h^*), \qquad
  S(c, h^*) = |\mathcal{N}(c) \cap \mathcal{N}(h^*)|,
\]
and leave the correct head's score unchanged,
$\hat{s}_{\mathrm{oracle}}(h^*) = \hat{s}(h^*)$.
Here $\hat{s}(\cdot)$ is the RotatE decoder score of the trained
RotatE + \sysname{} model,
and $\mathcal{N}$ is the training neighborhood defined in
App.~\ref{app:proofs}. The oracle penalizes each candidate by the size of its
neighborhood overlap with the correct answer, scaled by $\alpha$. We sweep
$\alpha \in \{1, 2, 5, 10, 20\}$; at $\alpha = 5$ the oracle score achieves
$0.9340$ head MRR on WikiKG2, the reference level reported in
Sec.~\ref{sec:exp-robustness}.

The oracle is not a deployable method, since it requires knowing $h^*$ to
compute $S(c, h^*)$. Its purpose is to estimate the head MRR reachable once
doppelganger confusion is resolved, which we find to be $0.9340$. The natural
way to close the gap to this level is query conditioning: two similar
entities always look similar if one looks only at the entities, but the
query itself can be used to tell them apart.

\paragraph{Stratification by overlap magnitude.}
Head MRR is not uniform across queries. Stratifying head queries by the
overlap in $(r, t)$ pairs between the highest-ranked wrong head and the
correct head reveals a non-monotonic pattern, shown in
Tab.~\ref{tab:overlap-stratification}.

\begin{table}[ht]
\centering
\caption{Test head queries on WikiKG2 stratified by overlap in $(r, t)$ pairs
between the highest-ranked wrong head and the correct head.}
\label{tab:overlap-stratification}
\small
\begin{tabular}{r|rrr}
\toprule
Overlap & $N$ queries & \% of total & Head MRR \\
\midrule
$0$       & 56{,}696   & $9.5\%$  & $0.564$ \\
$1$--$2$  & 323{,}760  & $54.1\%$ & $0.423$ \\
$3$--$5$  & 215{,}499  & $36.0\%$ & $0.494$ \\
$6$--$10$ & 2{,}567    & $0.4\%$  & $0.649$ \\
\bottomrule
\end{tabular}
\end{table}

The $1$--$2$ overlap bucket contains the majority of queries ($54.1\%$) and
has the lowest head MRR ($0.423$). These are entities that share just enough
structure to confuse the model but not enough to provide a clear
disambiguating signal. This bucket is where modest query-conditioned signal
would have maximum impact. The higher-overlap buckets have higher head MRR.
This is consistent with Proposition~A, which bounds the distance between two
representations by the number of training pairs the entities do \emph{not}
share rather than by their overlap: two entities with a large overlap can
still differ in many pairs. This stratification identifies where any
query-conditioned method would need to intervene most aggressively.

\section{Theoretical Analysis of the Doppelganger Effect}
\label{app:proofs}

\subsection{Setup and Proposition Statement}

For an entity $e \in \mathcal{E}$, define its outgoing training neighborhood
$\mathcal{N}(e) = \{(r, t) : (e, r, t) \in \mathcal{T}\}$ and its
bidirectional neighborhood
$\widetilde{\mathcal{N}}(e) = \mathcal{N}(e) \cup \{(r, h) : (h, r, e) \in \mathcal{T}\}$.
\sysname{} uses $\widetilde{\mathcal{N}}$ in practice. The proofs use
$\mathcal{N}$ for readability, and every statement holds verbatim with
$\widetilde{\mathcal{N}}$ in its place. For Lemma~3 and App.~\ref{app:wl},
the pairs in $\widetilde{\mathcal{N}}$ are taken with their direction
(equivalently, an incoming pair $(r, h)$ is written with the inverse relation
$r^{-1}$), since an entity's vector enters a loss term differently as a head
and as a tail.
For a candidate $c$ and the correct answer $h^*$, write
$S(c, h^*) = |\mathcal{N}(c) \cap \mathcal{N}(h^*)|$ and
$\Delta(c, h^*) = \mathcal{N}(c) \triangle \mathcal{N}(h^*)$, so that
$|\Delta(c, h^*)| = |\mathcal{N}(c)| + |\mathcal{N}(h^*)| - 2S(c, h^*)$.
A feature $f : \mathcal{E} \to \mathbb{R}^d$ is \emph{query-independent} if
$f(e)$ is computed without reference to any specific query.

\begin{quote}
\textbf{Proposition A (Query-independent ceiling).}
\label{prop:query-independent}
Let $f$ be a query-independent structural feature from one of three classes:
\emph{(i)} hash-based (Bloom filters);
\emph{(ii)} histogram-based (RelDist); or
\emph{(iii)} gradient-trained (TransE, DistMult, ComplEx, jointly learned
entity tables), under the assumptions of Lemma~3.
Then $\|f(c) - f(h^*)\| \le D_f(|\Delta(c, h^*)|)$ for a non-decreasing
function $D_f$ with $D_f(0) = 0$ that does not depend on the pair
$(c, h^*)$. The same holds for any concatenation of features from these
classes, with or without a subsequent Lipschitz map (e.g., a linear
projection or a ReLU MLP).
\end{quote}

\subsection{Proof of Proposition~A}

\subsubsection{Lemma 1: Hash-Based Features}

$\|f_{\mathrm{Bloom}}(c) - f_{\mathrm{Bloom}}(h^*)\|_1
\le k \cdot |\Delta(c, h^*)|$.

\textit{Proof.} The $\ell_1$ distance between two binary filters is the
number of bits set in exactly one of them. Without loss of generality, let
bit $b$ be set in $f_{\mathrm{Bloom}}(c)$ but not in
$f_{\mathrm{Bloom}}(h^*)$: some pair $\sigma \in \mathcal{N}(c)$ hashes to
$b$ under one of the $k$ functions, and no pair in $\mathcal{N}(h^*)$ does.
Hence $\sigma \notin \mathcal{N}(h^*)$, so $\sigma \in \Delta(c, h^*)$. Each
pair in $\Delta(c, h^*)$ sets at most $k$ bits, so at most
$k|\Delta(c, h^*)|$ bits differ. \hfill$\square$

\paragraph{Remark (MinHash).}
A MinHash signature (Sec.~\ref{sec:exp-robustness}) satisfies the same
conclusion in expectation. Coordinate $i$ differs between $c$ and $h^*$ only
if the pair minimizing $h_i$ over $\mathcal{N}(c) \cup \mathcal{N}(h^*)$
lies in $\Delta(c, h^*)$; for random hash functions this happens with
probability $|\Delta(c, h^*)| / |\mathcal{N}(c) \cup \mathcal{N}(h^*)|$, so
the expected number of differing coordinates is
$k|\Delta(c, h^*)| / |\mathcal{N}(c) \cup \mathcal{N}(h^*)|
\le k|\Delta(c, h^*)|$.

\subsubsection{Lemma 2: Histogram Features}

Let the histogram have bins $\{P_j\}$ that partition the possible $(r, x)$
pairs (for RelDist, one bin per relation,
$P_r = \{(r, x) : x \in \mathcal{E}\}$), with
$f_j(e) = |\mathcal{N}(e) \cap P_j|$. Then
$\|f_{\mathrm{hist}}(c) - f_{\mathrm{hist}}(h^*)\|_1 \le |\Delta(c, h^*)|$.

\textit{Proof.} For each bin,
$|f_j(c) - f_j(h^*)| \le |(\mathcal{N}(c) \triangle \mathcal{N}(h^*)) \cap P_j|$,
because pairs in both neighborhoods contribute equally to both counts.
Summing over the disjoint bins gives at most $|\Delta(c, h^*)|$.
\hfill$\square$

\subsubsection{Lemma 3: Gradient-Trained Features}

Let $f_{\mathrm{grad}}(e) = \theta_e^*$ be the entity vectors at a stationary
point $\theta^*$ of the training loss $L$. Fix all parameters other than
$\theta_e$ at $\theta^*$, and assume the following hold for $e \in \{c, h^*\}$,
with constants $G$ and $\mu$ common to all entity pairs considered.
\begin{itemize}
\item[(A1)] \emph{Per-entity decomposition.} The terms of $L$ that depend on
$\theta_e$ are
$F_e(\theta) = \sum_{\sigma \in \mathcal{N}(e)} \phi_\sigma(\theta) + \Psi(\theta)$,
where $\phi_\sigma$ is the loss term of training pair $\sigma = (r, t)$
viewed as a function of the entity's own vector, which is the same function
for every entity whose neighborhood contains $\sigma$, and $\Psi$ collects
terms that are the same for every entity (e.g., a per-entity regularizer, or
the expected negative-sampling term under uniform corruption).
\item[(A2)] \emph{Bounded gradients.} $\|\nabla \phi_\sigma\|_2 \le G$ on a
convex set $U$ containing $\theta_c^*$ and $\theta_{h^*}^*$.
\item[(A3)] \emph{Local strong convexity.} $F_e$ is $\mu$-strongly convex on $U$.
\end{itemize}
Then
\[
  \|f_{\mathrm{grad}}(c) - f_{\mathrm{grad}}(h^*)\|_2
  \le \frac{G}{\mu}\,|\Delta(c, h^*)|.
\]

\textit{Proof.} Stationarity gives $\nabla F_c(\theta_c^*) = 0$ and
$\nabla F_{h^*}(\theta_{h^*}^*) = 0$. By (A1), the shared pairs and $\Psi$
cancel in $F_c - F_{h^*}$, which therefore contains only the pairs in
$\Delta(c, h^*)$:
\[
  \nabla F_c(\theta_{h^*}^*)
  = \nabla F_c(\theta_{h^*}^*) - \nabla F_{h^*}(\theta_{h^*}^*)
  = \sum_{\sigma \in \mathcal{N}(c) \setminus \mathcal{N}(h^*)}
      \nabla\phi_\sigma(\theta_{h^*}^*)
  - \sum_{\sigma \in \mathcal{N}(h^*) \setminus \mathcal{N}(c)}
      \nabla\phi_\sigma(\theta_{h^*}^*),
\]
so $\|\nabla F_c(\theta_{h^*}^*)\|_2 \le G|\Delta(c, h^*)|$ by (A2). By (A3),
\[
  \mu\|\theta_{h^*}^* - \theta_c^*\|_2^2
  \le \big\langle \nabla F_c(\theta_{h^*}^*) - \nabla F_c(\theta_c^*),\,
      \theta_{h^*}^* - \theta_c^* \big\rangle
  \le \|\nabla F_c(\theta_{h^*}^*)\|_2\,\|\theta_{h^*}^* - \theta_c^*\|_2,
\]
hence $\|\theta_c^* - \theta_{h^*}^*\|_2
\le \|\nabla F_c(\theta_{h^*}^*)\|_2 / \mu
\le G|\Delta(c, h^*)| / \mu$. \hfill$\square$

\paragraph{Remarks on Lemma~3.}
\emph{Scaling with overlap.} If each $\phi_\sigma$ is $\mu_0$-strongly convex
on $U$ and $\Psi$ is $\mu_\Psi$-strongly convex, then
$\mu \ge \mu_\Psi + \mu_0|\mathcal{N}(c)|$; since
$|\mathcal{N}(c)| \ge S(c, h^*)$, the bound becomes
$G|\Delta(c, h^*)| / (\mu_\Psi + \mu_0 S(c, h^*))$, which tightens as the
shared overlap grows for a fixed number of unshared pairs.
\emph{On the assumptions.} KGE objectives are not globally convex, so (A3)
is an assumption about the region around the solution reached by training,
and the trained embeddings are idealized as a stationary point. Strong
convexity can come from the pair terms themselves (e.g., a squared-distance
objective) or from per-entity $\ell_2$ regularization. Where (A3) fails,
Lemma~3 gives no guarantee; Lemmas~1 and~2 hold unconditionally.

\subsubsection{Combination Under Concatenation and Lipschitz Maps}

Let $f_1, \ldots, f_n$ satisfy
$\|f_i(c) - f_i(h^*)\|_2 \le D_i(|\Delta(c, h^*)|)$ (the $\ell_1$ bounds of
Lemmas~1 and~2 also bound the $\ell_2$ distance). The concatenation
$f = [f_1; \ldots; f_n]$ satisfies
$\|f(c) - f(h^*)\|_2 \le \big(\sum_i D_i^2\big)^{1/2}$, which is
non-decreasing in $|\Delta(c, h^*)|$ and zero when $\Delta(c, h^*) = \emptyset$.
An $L$-Lipschitz map $M$ applied afterwards preserves this up to the factor
$L$:
$\|M(f(c)) - M(f(h^*))\|_2 \le L\big(\sum_i D_i^2\big)^{1/2}$. This covers
\sysname{}'s projection and fusion MLPs (a ReLU MLP is Lipschitz with
constant at most the product of its layers' operator norms; dropout is
inactive at evaluation) and the gate of Eq.~(2), which is linear in its
inputs.

\paragraph{Corollary A (score collapse).}
Let $z(e)$ be a representation covered by Proposition~A, with
$\|z(c) - z(h^*)\| \le D(|\Delta(c, h^*)|)$, and let the decoder score
$\hat{s}_q(e)$ of a query $q$ be $L_s$-Lipschitz in $z(e)$. Then, for every
query $q$,
\[
  |\hat{s}_q(c) - \hat{s}_q(h^*)| \le L_s\, D(|\Delta(c, h^*)|).
\]
For a RotatE head query $(?, r, t)$, the score
$\hat{s}_q(e) = \gamma - \|z(e) \circ \mathbf{r} - \mathbf{t}\|$ is
Lipschitz in $z(e)$, with constant $1$ in the norm used by the distance,
because rotation by $\mathbf{r}$ ($|r_i| = 1$) preserves the modulus of each
coordinate. Candidates whose neighborhoods nearly coincide with that of $h^*$
therefore receive nearly the same score as $h^*$ on every query, so no
representation covered by Proposition~A can reliably rank $h^*$ above them.
This is the sense in which Proposition~A implies a ceiling.

\paragraph{Implication.}
In the decoder-only configuration on WikiKG2, which has no domain-specific
features, every input to the decoder is covered by Proposition~A: the
structural priors, the jointly learned entity table, and their gated fusion.
App.~\ref{app:wl} shows that message passing over the training graph does
not separate entities with identical neighborhoods either. Separating such
entities requires conditioning on the query.

\subsection{Connection to Message Passing and 1-Weisfeiler--Leman}
\label{app:wl}

Consider a message-passing GNN in which each layer computes an entity's new
representation from its current representation and the multiset of messages
from its neighbors, $\{(r, h_x^{(l)}) : (r, x) \in \widetilde{\mathcal{N}}(e)\}$,
applied over full neighborhoods (as at inference). This covers relational
convolutions, mean aggregation, and attention, whose weights depend only on
these same inputs. If $\widetilde{\mathcal{N}}(c) = \widetilde{\mathcal{N}}(h^*)$
and $h_c^{(0)} = h_{h^*}^{(0)}$, then by induction every layer applies the
same function to the same arguments for both entities, so
$h_c^{(l)} = h_{h^*}^{(l)}$ at every depth. Since
$\widetilde{\mathcal{N}}(c) = \widetilde{\mathcal{N}}(h^*)$ implies
$\Delta(c, h^*) = \emptyset$, Proposition~A gives equal query-independent
inputs, and message passing therefore cannot separate the two entities.
This is the simplest instance of the 1-WL limitation of message-passing
GNNs~\cite{xu2018powerful}. Structural priors and message-passing GNNs thus
share the same limitation for entities with identical neighborhoods. Query
conditioning, as in labeling tricks~\cite{seal2} and
NBFNet~\cite{nbfnet}, changes the inputs per query and falls outside this
argument.
\section{Extended Related Work}
\label{app:related-extended}

This appendix expands on the classification of related work presented in
Sec.~\ref{sec:related-work}.

\paragraph{Scalable GNN training.}
Sampling-based methods reduce per-step computation by working
on subsets of the graph. GraphSAGE~\cite{graphSAGE} samples a
fixed-size neighborhood per node; FastGCN~\cite{fastgcn} uses
importance sampling at the layer level; ClusterGCN~\cite{clustergcn}
constructs mini-batches from graph partitions, and
GraphSAINT~\cite{graphsaint} from sampled subgraphs.
TIGER~\cite{tiger} extends inductive GNN
training to KG reasoning at scale through a structured
training pipeline. \sysname{} is complementary to all of
these: rather than reducing message-passing cost, it
precomputes structural features that remain informative
regardless of which subset is sampled at each step, directly
compensating for the structural information that sampling
discards. The Bloom filter and global TransE are computed once
over the full training graph and accessed in constant time
per entity.

\paragraph{Precomputed structural features.}
DeepWalk~\cite{deepwalk} and node2vec~\cite{node2vec} learn
node embeddings from random walks on homogeneous graphs.
Positional encodings for GNNs~\cite{positional_encoding}
augment node features with structural information about
position in the graph.
SIGN~\cite{sign} precomputes propagated features so that
training needs no message passing; it is a backbone rather than
a structural prior, and it composes with \sysname{}, which lifts
SIGN from 37.54\% to 40.46\% on OGB-MAG
(App.~\ref{app:nodeprop-extra-backbones}).
Bloom signatures~\cite{bloomsig} augment message-passing GNNs
with compact hashed encodings of node neighborhoods for scalable
link prediction on homogeneous graphs.
For heterogeneous KGs,
NodePiece~\cite{nodepiece} represents each entity by a fixed set
of anchor tokens and its relational context, whose embeddings are
learned jointly with the link-prediction objective.
\sysname{} differs from these in three ways. First,
its structural priors stay frozen across tasks, while
NodePiece learns its anchor embeddings for each downstream
objective.
Second, it combines \emph{both} local (Bloom) and
global (TransE) signals, while Bloom signatures~\cite{bloomsig}
and similar methods rely on a single feature type.
Third, it operates on top of the
learned embedding through a gated fusion rather than replacing
it, allowing the model to balance precomputed structural signal
against task-specific learning during training.

\paragraph{Local--global representation learning.}
Combining local and global information is a recurring idea in
other domains: time-series forecasting~\cite{ding2025global},
multimodal sentiment analysis~\cite{du2024hierarchical}, node
classification~\cite{eliasof2024global}, protein--ligand binding
affinity prediction~\cite{chen2025local}, and self-supervised
learning on molecular graphs~\cite{xu2021self}. In each case,
both representations are learned jointly with the training
objective on inputs that fit in memory. \sysname{} differs in that
its priors are frozen and computed once before any task training,
target heterogeneous KGs with hundreds of relation types, and are
reused unchanged across tasks and backbones.

\paragraph{Heterogeneous graph methods.}
Heterogeneous graph transformers such as HGT~\cite{hgt} apply
type-aware attention over meta-relations, and
PHGT~\cite{PHGT} extends this with poly-tokenization for
heterogeneous benchmark performance, achieving state-of-the-art
results on DBLP, ACM, and IMDb through a heavyweight
transformer architecture trained per-dataset. HOPE~\cite{hope}
and HGAMLP~\cite{hgamlp} are more recent approaches in the
same space, optimizing different aspects of heterogeneous
graph learning. \sysname{} reaches comparable or stronger
accuracy on the same benchmarks with a GraphSAGE backbone and
a single preprocessing recipe (Bloom + TransE) applied without
per-dataset tuning, supporting the paper's claim that
precomputed structural priors can substitute substantially for
purpose-built architectures.

\paragraph{Query-conditioned KG completion.}
NBFNet~\cite{nbfnet} and
A\textsuperscript{$\star$}Net~\cite{astartnet} compute query-specific
entity representations through path-based propagation.
More generally, labeling tricks~\cite{seal2}
condition node representations on the target link, which
provably separates links that independently computed node
representations cannot.
These methods escape the limit characterized by
Proposition~A, but at a per-query
cost: NBFNet exhausts GPU memory on WikiKG2 (as documented
in~\cite{astartnet}), and A\textsuperscript{$\star$}Net's accuracy
falls below the \sysname{} decoder despite its higher
representational capacity per query. \sysname{} occupies a
different point in the design space, trading per-query
expressiveness for the scalability of precomputed
query-independent features, a trade-off that proves favorable
on benchmarks where rare entities dominate the query
distribution.

\paragraph{Long-tail entity prediction.}
Popularity-agnostic evaluation~\cite{popularity} documents
that standard benchmarks under-weight rare-entity performance,
with embedding model accuracy degrading sharply as entity
popularity decreases. Most existing approaches address this
indirectly through training-time interventions such as
adjusted negative sampling or loss reweighting. \sysname{}
addresses long-tail entities directly: structural features
remain informative regardless of training frequency, and
Sec.~\ref{sec:exp-link-headline}'s stratified analysis
documents the benefit of capturing the long tail empirically across the bottom three
frequency buckets that account for $99\%$ of test queries.

\end{document}